\documentclass{article}

\PassOptionsToPackage{numbers, compress}{natbib}

\usepackage[eandd,preprint]{neurips_2026}

\usepackage[utf8]{inputenc} 
\usepackage[T1]{fontenc}    
\usepackage{hyperref}       
\usepackage{url}            
\usepackage{booktabs}       
\usepackage{amsfonts}       
\usepackage{nicefrac}       
\usepackage{microtype}      
\usepackage{xcolor}         
\usepackage[table]{xcolor}
\usepackage{amssymb}
\definecolor{lightgreen}{RGB}{240,250,240}   
\definecolor{green100}{RGB}{215,239,216}     
\definecolor{verylightgreen}{RGB}{247,252,247}
\definecolor{green80}{RGB}{228,245,229}      
\definecolor{verylightblue}{RGB}{250,253,255} 
\definecolor{lightblue1}{RGB}{230,240,255}   
\definecolor{lightblue2}{RGB}{208,230,255}   
\definecolor{lightblue3}{RGB}{184,217,255}   
\definecolor{lightblue4}{RGB}{214,236,252}   

\definecolor{checkgreen}{RGB}{0,140,60}
\definecolor{partialorange}{RGB}{220,140,0}
\definecolor{xmarkred}{RGB}{200,40,40}
\definecolor{groupgray}{RGB}{240,240,240}
\definecolor{ourshighlight}{RGB}{232,244,255}

\newcommand{\cmarkc}{\textcolor{checkgreen}{\ding{51}}}

\newcommand{\xmarkc}{\textcolor{xmarkred}{\ding{55}}}

\usepackage{pifont}  
\usepackage{booktabs}  

\usepackage{adjustbox}
\usepackage{amsmath}
\usepackage{multicol}
\usepackage{multirow}
\usepackage{refcount}

\usepackage{wrapfig}

\usepackage[most]{tcolorbox}
\tcbset{
  promptbox/.style={
    colback=gray!5,
    colframe=black!70,
    boxrule=0.5pt,
    arc=2pt,
    left=6pt,
    right=6pt,
    top=6pt,
    bottom=6pt,
    breakable
  }
}

\usepackage{caption}
\title{Can LLM Agents Be CFOs? Benchmarking Long-Horizon Resource Allocation in an Uncertain Enterprise Environment}

\author{
\small
\textbf{Yi Han\textsuperscript{1}},
\textbf{Yan Wang\textsuperscript{2}}\thanks{Correspondence to: wy2266336@gmail.com, lfqian94@gmail.com},
\textbf{Lingfei Qian\textsuperscript{2}}\footnotemark[1],
\textbf{Haohang Li\textsuperscript{3}},
\textbf{Yupeng Cao\textsuperscript{3}},
\textbf{Yueru He\textsuperscript{4}},
\\
\small
\textbf{Xueqing Peng\textsuperscript{2}},
\textbf{Nanhan Shen\textsuperscript{1}},
\textbf{Yitao Xu\textsuperscript{1}},
\textbf{Yankai Chen\textsuperscript{6,7}},
\textbf{Dongji Feng\textsuperscript{8}},
\\
\small
\textbf{Jimin Huang\textsuperscript{2,9}},
\textbf{Xue Liu\textsuperscript{6,7,10}},
\textbf{Jian-Yun Nie\textsuperscript{11}},
\textbf{Sophia Ananiadou\textsuperscript{9}}
\\[4pt]
\small
\textsuperscript{1}Georgia Institute of Technology,
\textsuperscript{2}The Fin AI,
\textsuperscript{3}Stevens Institute of Technology,
\\
\small
\textsuperscript{4}Columbia University,
\textsuperscript{5}George Mason University,
\textsuperscript{6}McGill University,
\\
\small
\textsuperscript{7}Mohamed bin Zayed University of Artificial Intelligence,
\\
\small
\textsuperscript{8}California State University, Monterey Bay,
\textsuperscript{9}University of Manchester,
\\
\small
\textsuperscript{10}Mila -- Quebec Artificial Intelligence Institute,
\textsuperscript{11}Université de Montréal
\\
}

\begin{document}

\maketitle

\begin{abstract}
Large language model (LLM) agents are increasingly tested on complex tasks, but their ability to allocate scarce resources over long horizons remains unclear. Unlike reactive tasks with immediate feedback, this setting requires agents to make binding commitments under partial observability, delayed consequences, hard resource budgets, and shifting dynamics. We introduce \textbf{\textsc{EnterpriseArena}}, a 132-month CFO simulator that evaluates long-horizon resource allocation under uncertainty in a FinTech lending firm. Agents must manage liquidity, close books, gather costly signals, and request equity or debt financing across changing macroeconomic regimes. The simulator is built from transformed firm-level financial data, anonymized business documents, decade-scale macroeconomic and industry signals, and expert-validated operating rules. Experiments across 23 LLMs and four agent frameworks show that current agents remain far from robust: only 15.4\% of trials survive the full horizon, larger models do not reliably outperform smaller ones, and failures cascade across observation, action timing, and capital sizing. These findings establish long-horizon resource allocation under uncertainty as a distinct capability gap for LLM agents. We have released our code \footnote{https://anonymous.4open.science/r/CFO-Env-F1B9}.

\end{abstract}

\section{Introduction}

Benchmarks for large language model (LLM) agents have rapidly expanded from web navigation~\cite{zhou2023webarena, xu-etal-2025-turkingbench} and software engineering~\cite{deng2025swebenchproaiagents, zhoufeaturebench} to tool use~\cite{guo2026mcp} and financial applications~\cite{li-etal-2025-investorbench, qian2026agents}. As LLMs~\cite{chang2024survey} evolve from passive assistants into agents that reason, plan, and act, these benchmarks test increasingly complex forms of competence. Yet one capability remains underexplored: \textbf{Can LLM agents allocate scarce resources under long-horizon uncertainty?}

We define this capability as an agent's ability to commit scarce, non-recoverable resources when their value cannot be immediately verified, and to sustain a coherent strategy across many such commitments as conditions evolve~\cite{dixit1994investment,puterman1990markov}. This setting differs from the largely reactive competence tested by most existing agent benchmarks~\cite{zhou2023webarena, xu-etal-2025-turkingbench,guo2026mcp,zhao2026amabenchevaluatinglonghorizonmemory,xu2025theagentcompanybenchmarkingllmagents,zhoufeaturebench, deng2025swebenchproaiagents}. Prior tasks often involve explicit action spaces, immediate feedback, and local errors that can be corrected in later attempts. By contrast, evaluating long-horizon resource allocation requires four structural properties to hold simultaneously: \textbf{hard resource budgets}, where both action and observation consume scarce capacity~\cite{altman2021constrained,hady2025multi}; \textbf{long horizons}, where reasoning and commitments must remain coherent across many decision steps~\cite{diamond2016liquidity}; \textbf{latent consequences}, where the true state and action outcomes are only partially revealed over time~\cite{kaelbling1998planning}; and \textbf{non-stationary dynamics}, where transitions depend on both exogenous shifts and the agent's prior actions~\cite{sutton1998reinforcement,hady2025multi} (more details are listed in Table~\ref{tab:benchmark_comparison} in Appendix~\ref{compare}). Formally, this corresponds to a constrained partially observable Markov decision process with latent rewards under non-stationary dynamics~\cite{puterman1990markov}.

Existing financial agent benchmarks come closest to this setting, but each misses part of the structure. Signal-response benchmarks~\cite{qian2026agents, fan2025aitraderbenchmarkingautonomousagents, chen2025stockbench} test reactions to market signals, but trades are often reversible, capacity is effectively unbounded, and feedback is frequent. Judgment-oriented benchmarks~\cite{li2025investorbench, bigeard2025financeagentbenchmarkbenchmarking} evaluate investment recommendations, but recommendations remain static outputs rather than environment-level actions. Workflow benchmarks~\cite{zeng2025fingaiachinesebenchmarkai} assess multi-step financial reasoning, but usually in fixed scenarios whose dynamics do not evolve in response to the agent. None places the agent in a setting where scarce capacity must be allocated across long horizons, consequences emerge only over time, and dynamics shift across regimes.

\begin{figure*}[t]
    \centering
    \includegraphics[width=0.9\linewidth]{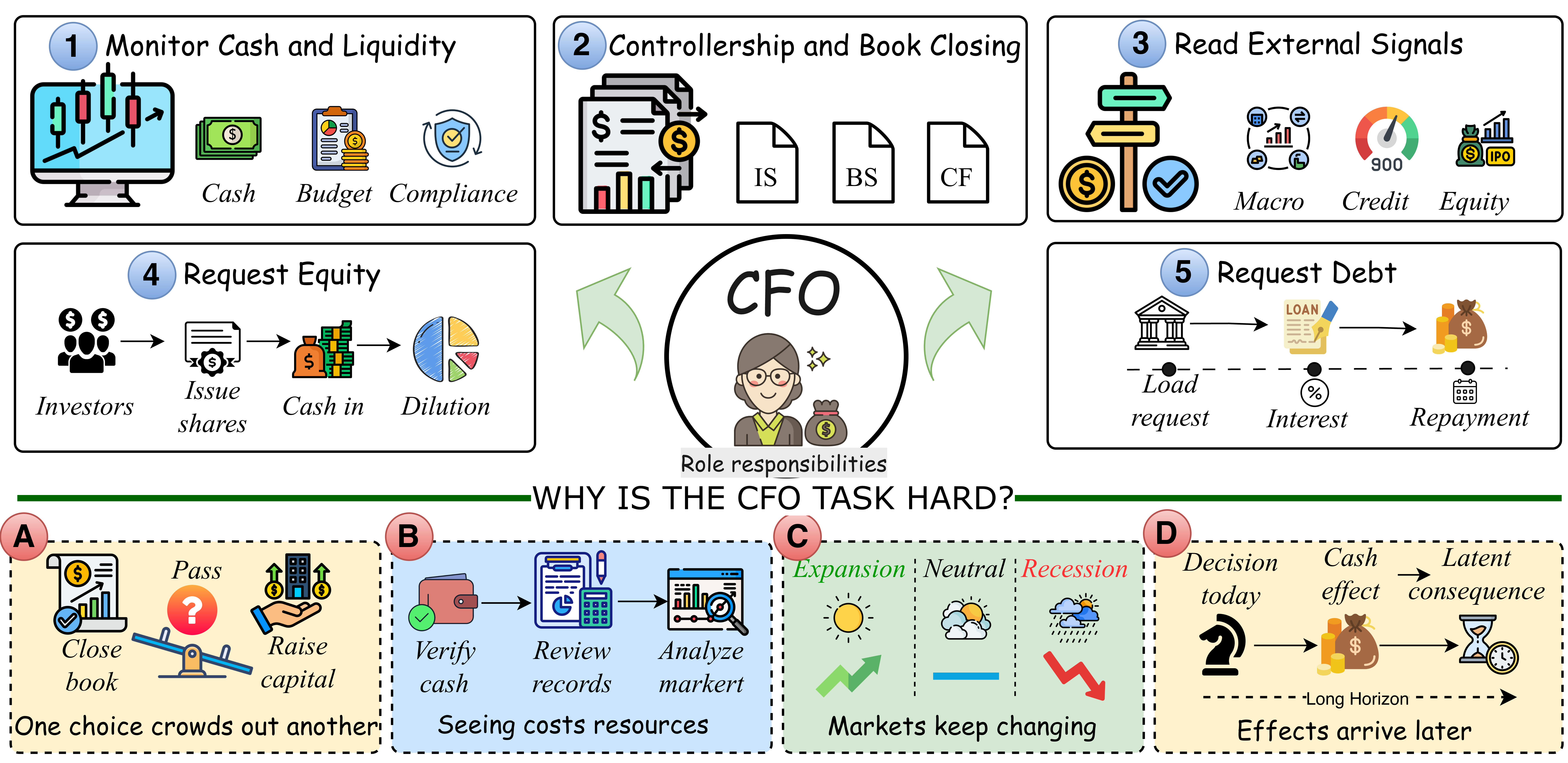}
    \caption{CFO functionality in an organization, illustrating how the setting instantiates the four structural conditions required for long-horizon resource allocation under uncertainty. \emph{Top:} five core financial functions. \emph{Bottom:} four characteristics (A--D) that make the CFO setting a natural stress test for this target capability.}
    \label{fig:cfo}
\end{figure*}

To study it concretely, we use the role of a Chief Financial Officer (CFO) at a FinTech lending firm as a high-resolution testbed rather than as a general claim about CFO work. This setting naturally exposes the target challenge: the CFO must decide whether to close books, raise capital, or hold (Fig.~\ref{fig:cfo}A); observing the firm's true financial state requires costly organization investigation rather than free signal (Fig.~\ref{fig:cfo}B); macroeconomic conditions shift across expansion, neutral, and recession phases (Fig.~\ref{fig:cfo}C); and decisions take effect only after delays, with consequences unfolding over an extended horizon (Fig.~\ref{fig:cfo}D). We instantiate this setting as \textsc{EnterpriseArena}, a 132-month enterprise simulator built on transformed firm-level financial data, anonymized business documents, decade-scale macroeconomic and industry signals, and operating rules validated by enterprise finance experts. The environment exposes a unified interface in which agents must choose among capacity-consuming actions and costly observation tools while managing liquidity, fundraising, debt burden, and operational uncertainty across multiple macroeconomic regimes.

We evaluate four state-of-the-art agentic frameworks: ReAct~\cite{yao2022react}, Claude Code\footnote{\url{https://code.claude.com/docs/en/overview}\label{cc}}, Codex\footnote{\url{https://developers.openai.com/codex}\label{cx}}, and OpenClaw\footnote{\url{https://docs.openclaw.ai/}\label{oc}}. Under ReAct, we test 23 recent backbone LLMs ranging from below 10B parameters to several hundred billion parameters; the other frameworks are evaluated with their native backbones. The results reveal a substantial capability gap. Only 15.4\% of trials survive the full 132 months, and 12 of the 23 ReAct backbones never survive once. Model scale does not reliably predict success: an 8B Llama-3.1 model achieves a terminal score of \$30.6M, nearly 2$\times$ that of the 397B Qwen3.5-MoE (\$16.0M), and higher than all seven closed-source frontier backbones we test. Stronger frameworks improve performance but remain far from human experts: the best configuration, Codex CLI with GPT-5.5, reaches \$34.7M, only about 7\% of the human expert baseline (\$476.7M). We trace these failures to three hierarchically linked breakdowns in the allocation process: agents allocate observation capacity toward internal state rather than external signals, mis-time their first fundraising action until cash has already begun to decline, and under-size their capital requests as a consequence of insufficient context.

We make three contributions:
(1) We formalize \emph{long-horizon resource allocation under uncertainty} as a distinct evaluation target for LLM agents.
(2) We present \textsc{EnterpriseArena}, a 132-month CFO simulator for benchmarking this capability.
(3) We provide empirical evidence that current agents remain far from robust, with low survival rates and recurring strategic failures across 23 LLMs and four agent frameworks.

\section{Related Work}

\noindent{\textbf{Financial Agent Benchmarks.}}
Recent work has increasingly explored LLMs for financial modeling, reasoning, and auditing~\cite{qian2025fino1,wang2025rkefino1regulationknowledgeenhancedlarge,huang2025openfinllmsopenmultimodallarge,wang2026finauditingfinancialtaxonomystructuredmultidocument}. Building on these advances, financial agent benchmarks evaluate tasks such as trading, investment recommendation, financial analysis, and tool-augmented reasoning~\cite{li2025investorbench, qian2026agents, fan2025aitraderbenchmarkingautonomousagents, chen2025stockbench, bigeard2025financeagentbenchmarkbenchmarking, zeng2025fingaiachinesebenchmarkai}. However, these settings primarily test market-facing decisions or static financial workflows. They do not place the agent in the role of an internal corporate decision-maker who must manage an enterprise over time, allocate scarce capacity, and bear the delayed consequences of prior commitments.

\noindent{\textbf{Agent Environment Benchmarks.}}
Beyond finance, LLM agents have been evaluated in interactive environments involving web navigation, tool use, software engineering, workplace simulation, memory, safety, and continual learning~\cite{liu2025agentbenchevaluatingllmsagents, zhou2023webarena, xu-etal-2025-turkingbench, deng2025swebenchproaiagents, zhoufeaturebench, guo2026mcp, xu2025theagentcompanybenchmarkingllmagents, vishwakarma2025llmshelpworksandbox, zhao2026amabenchevaluatinglonghorizonmemory, tur2025safearena, zheng2025lifelongagentbench, shridhar2020alfworld}. These benchmarks capture important aspects of sequential interaction, tool use, and long-horizon behavior. In most cases, however, the agent either completes tasks in environments with explicit feedback, operates over pre-recorded trajectories, or pursues goals where errors remain local and recoverable. \textsc{EnterpriseArena} instead evaluates closed-loop enterprise decision-making, where each action consumes scarce capacity, observations are costly, consequences are delayed, and prior commitments reshape future states.


\section{EnterpriseArena}
\label{sec:enterprisearena}



\subsection{Task Formulation}
\begin{figure*}
    \centering
    \includegraphics[width=0.95\linewidth]{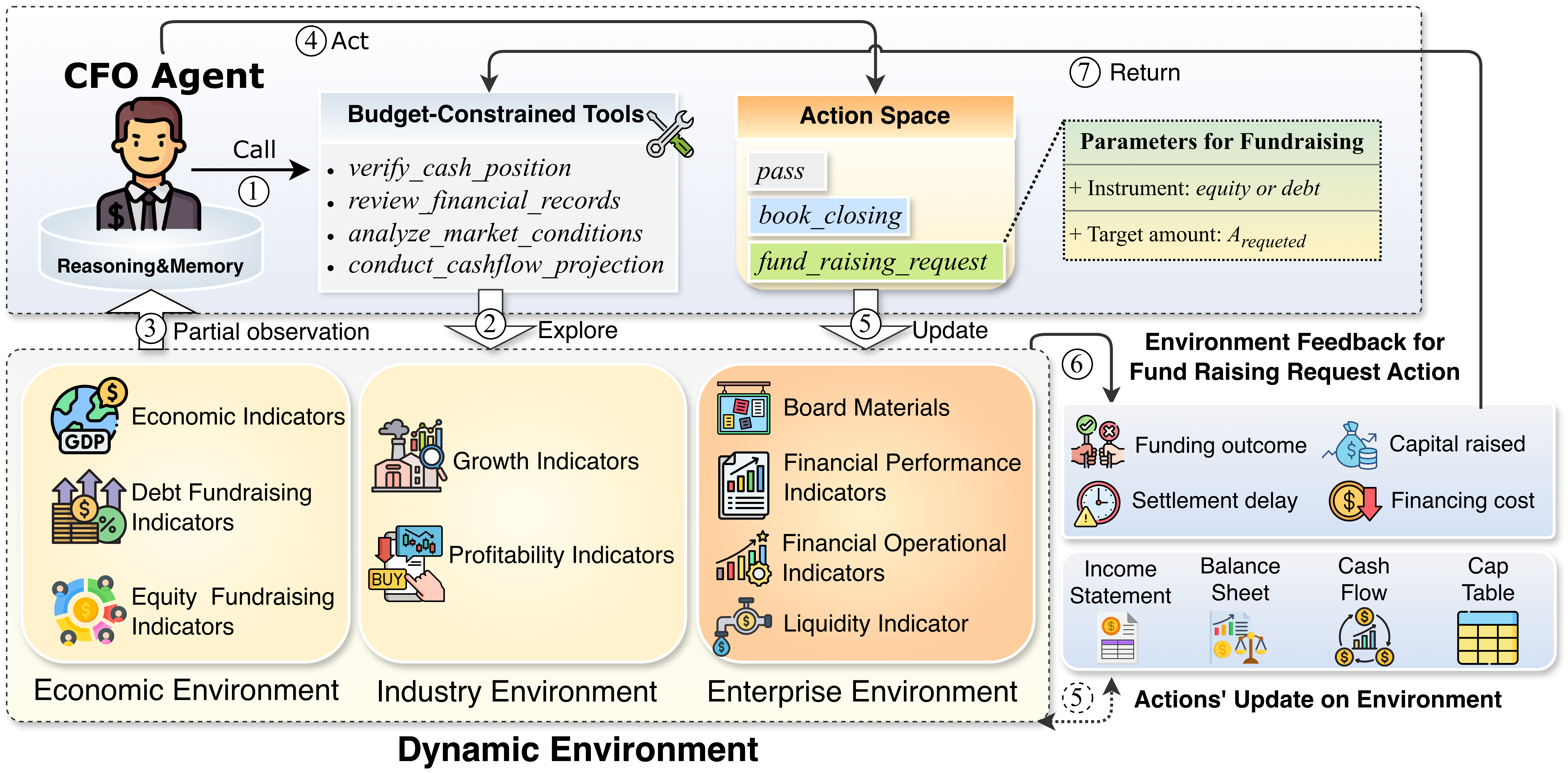}
    \caption{Overview for \textsc{EnterpriseArena} Benchmark.}
    \label{fig:framework}
\end{figure*}

We formulate \textsc{EnterpriseArena} as a long-horizon agentic decision-making problem for resource allocation under uncertainty in an enterprise financial environment, as shown in Figure~\ref{fig:framework}. The agent acts as the CFO of a simulated enterprise, making sequential monthly decisions over $T$ timesteps. 
Its primary objective is survival: the company's cash balance must remain non-negative at every timestep, and violating this constraint terminates the episode with a score of zero. Subject to survival, the agent aims to maximize terminal enterprise valuation at the final timestep, reflecting long-term business growth.

The central design principle of \textsc{EnterpriseArena} is \emph{to induce organizational-level trade-offs}. In real enterprises, activities such as reconciling financial records and raising capital require limited teams, time, and infrastructure, and therefore cannot be pursued freely or simultaneously.\cite{campello2024corporate}
\textsc{EnterpriseArena} reflects this constraint throughout the task design. The environment evolves through stochastic dynamics (Section~\ref{sec:env}), the agent can only access the state through budget-constrained tools (Section~\ref{sec:obs}), and each action requires trading off between improving visibility through reconciliation and strengthening liquidity through capital acquisition (Section~\ref{sec:action}).

\subsection{Dynamic Layered Environment}
\label{sec:env}

The environment models two layers of dynamics: internal operations that drive the firm's monthly financial activity, and external conditions that shape the broader context.

\noindent{\textbf{Internal enterprise dynamics.}}
The enterprise state includes the firm's financial position, user base, contracts, and accumulated organizational records, initialized with financial statements, governance documents, an initial cash balance, and an initial user count (details in Appendix~\ref{app_obs_section}). At each timestep, the state transition is governed by $n$ operational indicators $\{x_1, \ldots, x_n\}$ controlling different dimensions of the firm's activity, such as revenue generation and expenditure~\cite{kao2025business}. To simulate real-world unpredictability, each indicator is independently perturbed:
\begin{equation}
    x_i' = x_i + \epsilon_i, \quad \epsilon_i \sim \mathcal{N}(0, \sigma_i^2),
\end{equation}
where each $\sigma_i$ is calibrated to reflect that indicator's real-world volatility (details in Appendix~\ref{dynamic}). Because multiple indicators vary simultaneously, the agent cannot rely on a single signal to track the enterprise's trajectory; it must integrate noisy observations across dimensions. Moreover, the consolidated financial position, actual profitability and outstanding obligations, is not directly observable and requires a formal reconciliation action, which consumes the agent's only action slot for that timestep.

\noindent{\textbf{External economic and industry dynamics.}}
The enterprise is also shaped by macroeconomic indicators (e.g., GDP growth, interest rates) and industry-level metrics (e.g., sector margins, user growth rates) beyond the agent's control (details in Appendix~\ref{app_obs_section}). These indicators follow a fixed trajectory derived from anonymized real-world historical data spanning multiple economic phases, including expansion, neutral, and recession periods~\cite{giampaoli2024business}. Unlike internal dynamics, this external trajectory is deterministic and exogenous, but unseen by the agent, it must infer the current regime from observed signals rather than being told which phase the economy is in. These external signals affect both enterprise state transitions and action outcomes. For example, fundraising success depends on market conditions at the time of the attempt (Section~\ref{sec:action}). See Appendix~\ref{dynamic} for details.
\subsection{Information Acquisition via Organizational Tools}
\label{sec:obs}

In real enterprises, the full organizational state is distributed across separate systems and teams~\cite{balaha2025analytical}, and visibility requires specific organizational operations. The agent cannot directly access the full enterprise state; instead, it must invoke staff operations to obtain partial views, each incurring organizational effort. Four tools are available:
\textbf{\textit{(1) verify\_cash\_position}}: returns the current cash balance as a single scalar, with no breakdown of what drove the number.
\textbf{\textit{(2) review\_financial\_records}}: compiles historical internal documents within $[0, t]$; structured reports may lag depending on how recently the agent has reconciled.
\textbf{\textit{(3) analyze\_market\_conditions}}: gathers historical external indicators within $[0, t]$, without forecasting.
\textbf{\textit{(4) conduct\_cashflow\_projection}}: builds a forward-looking cash flow model based on agent-provided assumptions; output quality depends entirely on input quality.
Each tool reveals only one slice of the state, and the quality of results depends on how recently the agent has performed a reconciliation action~\cite{dee2025critical} (\textbf{book\_closing} in Section~\ref{sec:action}). Since reconciliation consumes the agent's only action slot for the period, improving observation quality comes at the direct cost of forgoing other actions.

Each tool call corresponds to a CFO team's real-world activity that consumes organizational staff resources, team coordination capacity, and time.
Therefore, we constrain the agent to at most 20 tool calls per timestep, forcing it to prioritize information under resource constraints.

\subsection{Trade-off Action with Environment Interaction}
\label{sec:action}

At each timestep, the agent executes exactly one action: \textbf{book\_closing}, \textbf{fund\_raising\_request}, or \textbf{pass}. Only one can be selected per period, creating a core trade-off: reconciling improves visibility but forgoes capital acquisition, while fundraising strengthens liquidity but may be poorly timed without an up-to-date view. Because consequences are delayed and irreversible, the agent must anticipate future risks based on company status and commit its limited action capacity proactively, rather than waiting until a need becomes urgent.

\noindent{\textbf{book\_closing.}}
It triggers reconciliation, the environment consolidates all accumulated records and produces ground-truth financial statements (income statement, balance sheet, cash flow statement) up to timestep $t$, which become immediately available through the agent's tools. This is the only way to obtain an accurate view of the enterprise's true state~\cite{dee2025critical}. Without it, the agent relies on raw signals and outdated reports.

\noindent{\textbf{fund\_raising\_request.}}
Requests external capital by specifying an instrument type (equity or debt) and target amount $A_{\text{requested}}$. Debt introduces increases of future cash outflows~\cite{dainelli2024financial}; equity introduces no recurring costs~\cite{liu2023does}. The environment determines feedback along four dimensions (details are listed in Appendix~\ref{app:fundraising}): 
\textit{(1) Funding outcome:} success or failure, sampled from $\text{Bernoulli}(p_{\text{adj}})$ where $p_{\text{adj}} = p_{\text{macro}} \times m_{\text{company}}$. Here $p_{\text{macro}} \in [0,1]$ is a base rate from external market conditions and $m_{\text{company}} \in [0,1]$ is a penalty based on the enterprise's state; equity becomes harder with each successful round, debt becomes harder as leverage grows.
\textit{(2) Capital raised:} $A_{\text{actual}} = f \cdot A_{\text{requested}}$, where $f \sim \mathcal{U}(0.7,\, 1.0)$.
\textit{(3) Settlement delay:} funds arrive after $d \sim \mathcal{U}_{\mathbb{Z}}(1,\, 6)$ months.
\textit{(4) Contract cost (debt only):} interest rate determined by market conditions at settlement ($t{+}d$), unknown at request time.
On success, $A_{\text{actual}}$ is added to cash at $t + d$, and debt instruments introduce recurring interest obligations that alter future transition dynamics.

\noindent{\textbf{pass.}}
The agent takes no action and the environment advances by one month. This may be appropriate after a recent reconciliation when market conditions are unfavorable for fundraising.

\subsection{Dataset Curation and Construction}
\label{sec:data_construction}

\noindent{\textbf{Foundational Data Collection.}}
The benchmark requires a domain where resource allocation under uncertainty arises naturally rather than being artificially imposed. We select a FinTech consumer lending company because this domain exhibits all four structural properties organically: high-frequency loan originations and repayments produce continuous cash flows over a long horizon; the capital-intensive business model demands recurrent fundraising under exclusive organizational capacity; sensitivity to interest rates and credit cycles creates non-stationary dynamics; and multi-party due diligence in fundraising introduces real settlement delays. We collect 16 types of data across three layers~\cite{Xue2022ResearchOT,azevedo2021earnings}: firm-level financials from public filings that initialize the enterprise state at $t=0$; industry-level metrics\footnote{\url{https://www.mckinsey.com/industries/financial-services/our-insights/fintechs-a-new-paradigm-of-growth}; \url{https://www.verifiedmarketresearch.com/services-industry/}} that provide sector benchmarks across the 132-month horizon; and macroeconomic indicators\footnote{\url{https://openstax.org/books/principles-finance/pages/1-3-importance-of-data-and-technology}}~\cite{bok2018macroeconomic} that drive external dynamics including credit conditions and capital market cycles. Details are in Appendix~\ref{app:dataset}.


\noindent{\textbf{Data Anonymization and Stochastic Noise.}}
To ensure that agent performance reflects genuine allocation reasoning rather than memorized historical knowledge, all identifiable information is removed: enterprises are labeled ``Company XYZ,'' company-specific details are redacted, and calendar dates are replaced with anonymized labels (e.g., ``Jan 2xx0'') so that agents cannot exploit known events such as COVID-19 or specific rate-hike cycles. The underlying economic dynamics and regime transitions are fully preserved (see Appendix~\ref{app_obs_section} for details). Stochastic perturbations (Table~\ref{tab:env-dynamics} in Appendix~\ref{app:fundraising}) further introduce per-timestep variability, ensuring that identical strategies cannot succeed through deterministic replay.

\noindent{\textbf{Industry-Guided Business Rules.}}
The backend accrual-based and cash-based ledgers tracking that is used to generate financial statements is guided by accounting standards (GAAP/ASC)~\cite{securities2008topic,toerner2009guide} and industry practice~\cite{scott2015financial,graham2012research} to reflect real-world financial timing lag challenges in the evolving environment (details in Appendix \ref{app:fundraising}). Human experts also verified select trajectories consistent with standard accounting principles~\cite{securities2008topic,toerner2009guide} with details in Appendix \ref{human-valid}.

Fundraising results dynamics (details in Appendix \ref{dynamic}) are based on market evidence\footnote{\url{https://www.sweetstudy.com/questions/week1-19965789}}, academic research, and industry reports~\cite{cassar2007cash}, where approval, amount, and cost depend on macroeconomic conditions and firm-specific characteristics.

\subsection{Evaluation Metrics}
\label{sec:eval}

We evaluate agent performance along two complementary dimensions: whether the agent can keep the enterprise alive, and how effectively it grows the enterprise over the full horizon.
 \textbf{(1) Survival.}
The agent must maintain a non-negative cash balance at every timestep. If $\text{Cash}_t < 0$ for any $t$, the episode terminates immediately with a score of $0$. This binary constraint serves as the primary test of the agent's ability to manage short-term liquidity risk under uncertainty.
\textbf{(2) Terminal valuation score.}
For episodes that survive, we measure the agent's ability to grow the enterprise with minimal resources, using a terminal valuation score:
\begin{equation}
    \text{Score}_T = \text{Rev}_T \times m + \text{Cash}_T - \lambda \cdot N_{\text{tools}},
\end{equation}
where $\text{Rev}_T$ is the trailing-twelve-months revenue at the final timestep, $m=5$ is a fixed valuation multiple\footnote{This multiple is calibrated using typical revenue multiples for FinTech companies; see 
\href{https://firstpagesage.com/business/fintech-valuation-multiples/}{source} for the corresponding empirical distribution.}, $\text{Cash}_T$ is the remaining cash balance, $N_{\text{tools}}$ is the total number of tool calls across the episode, and $\lambda=5{,}000$ is a penalty coefficient in consultation with enterprise experts. The first term reflects enterprise growth through a standard revenue-based valuation. The second term rewards prudent cash management. The third term penalizes excessive tool usage, since each tool call consumes organizational resources.

\section{Experiments and Results}


\subsection{Experiment Settings}
\label{sec:exp}
 
\noindent{\textbf{Environment configuration.}}
The simulated enterprise is a consumer lending company initialized with \$15M in cash, 5,000 borrowers, an average loan size of \$10K, zero debt, and 10.5M equity shares outstanding at \$10/share. Each episode spans $T=132$ timesteps, corresponding to an 11-year horizon with monthly updates. This length is chosen to cover multiple economic cycles (expansion, neutral, and recession phases), requiring the agent to adapt its strategy across varying conditions rather than optimizing for a single regime. All agents receive the same system prompt and environment configuration; stochastic noise is applied independently across trials.
 
\noindent{\textbf{Evaluated agents and backbone models.}}
We evaluate two complementary axes. The first is the \emph{backbone-model} axis: we hold the agent framework fixed at ReAct~\cite{yao2022react} and vary the underlying LLM, isolating how much of the gap on \textsc{EnterpriseArena} comes from the model itself rather than from framework engineering. 
The second is the agent \emph{framework} axis: we evaluate three state-of-the-art agent frameworks, Claude Code\hyperref[cc]{\textsuperscript{\getrefnumber{cc}}}, Codex\hyperref[cx]{\textsuperscript{\getrefnumber{cx}}}, and OpenClaw\hyperref[oc]{\textsuperscript{\getrefnumber{oc}}}, each paired with its native model. These frameworks are designed for long-horizon agentic work and use richer scaffolding than ReAct.

\emph{For the ReAct agent}, we powered it with 23 LLMs spanning four categories, closed-source models, large open-source models, medium-scale models and small-scale models from various LLM families. 
\emph{For the agent framework axis}, we pair each framework with its native frontier backbone unmodified: Claude Opus 4.7 for Claude Code, GPT-5.5 for Codex, and DeepSeek-V4 for OpenClaw. These three backbones are identical to those in our ReAct evaluation, enabling per-backbone comparisons that isolate the framework's contribution from the model's. Per-model details are in Appendix~\ref{app:models}.



\begin{table*}[t]
\centering
\footnotesize
\setlength{\tabcolsep}{2pt}
\renewcommand{\arraystretch}{1}
\resizebox{\textwidth}{!}{
\begin{tabular}{l|ccc|ccc|ccc}
\toprule
\textbf{Models}
& \multicolumn{3}{c|}{\textbf{Overall ($\uparrow$)}}
& \multicolumn{3}{c|}{\textbf{Multi-Crisis Survival ($\uparrow$)}}
& \multicolumn{3}{c}{\textbf{Agent Performance ($\uparrow$)}} \\
& Full Surv.\% & Avg. Mon. & Score (\$M)
& 1st Crisis & 2nd Crisis & 3rd Crisis
& Tools/Mo & Actions & Raised (\$M) \\
\midrule
Human
& \cellcolor{lightgreen}\textbf{60\%} 
& 92$_{\textcolor{gray}{\pm53}}$ 
& \textbf{476.7$_{\textcolor{gray}{\pm899.1}}$}
& \cellcolor{green100}100\% 
& \cellcolor{lightgreen}60\% 
& \cellcolor{lightgreen}60\%
& \cellcolor{verylightblue}0.58 
& \cellcolor{lightblue4}68.0 
& \textbf{1238.7$_{\textcolor{gray}{\pm2518.7}}$} \\
\midrule
\multicolumn{10}{c}{\textit{ReAct Agent Framework}} \\
\midrule
\multicolumn{10}{c}{\textit{Closed-source LLMs}} \\
\midrule
GPT-5.5
& 0\% & 49$_{\textcolor{gray}{\pm15}}$ & 0.0$_{\textcolor{gray}{\pm0.0}}$
& \cellcolor{green100}100\% & \cellcolor{lightgreen}60\% & 0\%
& \cellcolor{lightblue1}1.64 & \cellcolor{lightblue2}17.2 & 58.4$_{\textcolor{gray}{\pm47.3}}$ \\
GPT-5.4
& 0\% & 43$_{\textcolor{gray}{\pm14}}$ & 0.0$_{\textcolor{gray}{\pm0.0}}$
& \cellcolor{green100}100\% & 40\% & 0\%
& \cellcolor{verylightblue}0.76 & \cellcolor{verylightblue}0.4 & 1.7$_{\textcolor{gray}{\pm3.9}}$ \\
Gemini 3.1 Pro
& 20\% & 58$_{\textcolor{gray}{\pm43}}$ & 10.4$_{\textcolor{gray}{\pm23.3}}$
& \cellcolor{green100}100\% & 40\% & 20\%
& \cellcolor{lightblue2}2.46 & \cellcolor{lightblue2}13.2 & 41.4$_{\textcolor{gray}{\pm41.7}}$ \\
Claude Opus 4.7
& 20\% & 90$_{\textcolor{gray}{\pm42}}$ & 10.2$_{\textcolor{gray}{\pm23.0}}$
& \cellcolor{green100}100\% & \cellcolor{green80}80\% & \cellcolor{lightgreen}60\%
& \cellcolor{lightblue2}3.37 & \cellcolor{lightblue2}17.6 & \underline{73.4$_{\textcolor{gray}{\pm35.7}}$} \\
Claude-haiku
& 20\% & 63$_{\textcolor{gray}{\pm40}}$ & 20.0$_{\textcolor{gray}{\pm44.8}}$
& \cellcolor{green100}100\% & 60\% & 20\%
& \cellcolor{lightblue2}3.14 & \cellcolor{lightblue2}20.6 & 31.3$_{\textcolor{gray}{\pm44.8}}$ \\
Grok-4.20
& 0\% & 38$_{\textcolor{gray}{\pm12}}$ & 0.0$_{\textcolor{gray}{\pm0.0}}$
& \cellcolor{green100}100\% & 20\% & 0\%
& \cellcolor{verylightblue}0.26 & \cellcolor{verylightblue}1.2 & 0.0$_{\textcolor{gray}{\pm0.0}}$ \\
Grok-4.3
& \cellcolor{verylightgreen}\underline{40\%} 
& \underline{94$_{\textcolor{gray}{\pm45}}$} 
& \underline{25.4$_{\textcolor{gray}{\pm36.4}}$}
& \cellcolor{green100}100\% 
& \cellcolor{green80}80\% 
& \cellcolor{lightgreen}60\%
& \cellcolor{lightblue3}\underline{6.34} 
& \cellcolor{lightblue3}\underline{30.6} 
& 30.7$_{\textcolor{gray}{\pm19.7}}$ \\
\midrule
\multicolumn{10}{c}{\textit{Open-source LLMs (Large)}} \\
\midrule
GLM-5.1
& 0\% & 39$_{\textcolor{gray}{\pm11}}$ & 0.0$_{\textcolor{gray}{\pm0.0}}$
& \cellcolor{green100}100\% & 20\% & 0\%
& \cellcolor{lightblue3}7.41 & \cellcolor{lightblue1}5.8 & 26.9$_{\textcolor{gray}{\pm13.2}}$ \\
GLM-5
& 20\% & 69$_{\textcolor{gray}{\pm36}}$ & 29.2$_{\textcolor{gray}{\pm65.2}}$
& \cellcolor{green100}100\% & \cellcolor{green80}80\% & 20\%
& \cellcolor{lightblue2}3.26 & \cellcolor{lightblue2}11.6 & \underline{73.0$_{\textcolor{gray}{\pm52.5}}$} \\
Qwen3.5-397B-A17B
& 20\% & 69$_{\textcolor{gray}{\pm37}}$ & 16.0$_{\textcolor{gray}{\pm35.8}}$
& \cellcolor{green100}100\% & \cellcolor{green80}80\% & 20\%
& \cellcolor{lightblue2}3.86 & \cellcolor{verylightblue}4.4 & 25.9$_{\textcolor{gray}{\pm20.5}}$ \\
DeepSeek-V4
& \cellcolor{lightgreen}\underline{\textbf{60\%}} 
& \underline{97$_{\textcolor{gray}{\pm48}}$} 
& \underline{40.1$_{\textcolor{gray}{\pm37.3}}$}
& \cellcolor{green100}100\% 
& \cellcolor{green80}80\% 
& \cellcolor{lightgreen}60\%
& \cellcolor{lightblue3}\underline{\textbf{16.21}} 
& \cellcolor{lightblue3}24.0 
& 62.1$_{\textcolor{gray}{\pm39.5}}$ \\
DeepSeek-V3.1
& 0\% & 43$_{\textcolor{gray}{\pm15}}$ & 0.0$_{\textcolor{gray}{\pm0.0}}$
& \cellcolor{green100}100\% & 40\% & 0\%
& \cellcolor{lightblue2}4.85 & \cellcolor{lightblue1}7.0 & 5.5$_{\textcolor{gray}{\pm8.5}}$ \\
MiniMax-M2.7
& 20\% & 74$_{\textcolor{gray}{\pm32}}$ & 14.7$_{\textcolor{gray}{\pm33.0}}$
& \cellcolor{green100}100\% & \cellcolor{green100}100\% & 20\%
& \cellcolor{lightblue3}6.30 & \cellcolor{lightblue3}\underline{34.2} & 47.6$_{\textcolor{gray}{\pm28.0}}$ \\
Llama-3.3-70B-Instruct
& 0\% & 38$_{\textcolor{gray}{\pm11}}$ & 0.0$_{\textcolor{gray}{\pm0.0}}$
& \cellcolor{green100}100\% & 20\% & 0\%
& \cellcolor{lightblue2}4.59 & \cellcolor{verylightblue}2.6 & 0.0$_{\textcolor{gray}{\pm0.0}}$ \\
\midrule
\multicolumn{10}{c}{\textit{Open-source LLMs (Medium)}} \\
\midrule
Mistral-Small-24B-Instruct
& 0\% & \underline{50$_{\textcolor{gray}{\pm15}}$} & 0.0$_{\textcolor{gray}{\pm0.0}}$
& \cellcolor{green100}100\% & 60\% & 20\%
& \cellcolor{lightblue1}1.46 & \cellcolor{lightblue2}13.0 & 9.1$_{\textcolor{gray}{\pm9.3}}$ \\
Mixtral-8x7B-Instruct-
& 0\% & \underline{50$_{\textcolor{gray}{\pm15}}$} & 0.0$_{\textcolor{gray}{\pm0.0}}$
& \cellcolor{green100}100\% & 60\% & 0\%
& \cellcolor{verylightblue}0.56 & \cellcolor{lightblue1}5.4 & \underline{11.4$_{\textcolor{gray}{\pm7.1}}$} \\
Gemma-4-31B
& 0\% & 32$_{\textcolor{gray}{\pm1}}$ & 0.0$_{\textcolor{gray}{\pm0.0}}$
& \cellcolor{green100}100\% & 0\% & 0\%
& \cellcolor{lightblue2}2.22 & \cellcolor{verylightblue}4.8 & 0.0$_{\textcolor{gray}{\pm0.0}}$ \\
Qwen3.5-35B-A3B
& 0\% & 33$_{\textcolor{gray}{\pm1}}$ & 0.0$_{\textcolor{gray}{\pm0.0}}$
& \cellcolor{green100}100\% & 0\% & 0\%
& \cellcolor{lightblue2}\underline{2.79} & \cellcolor{lightblue2}\underline{14.2} & 8.0$_{\textcolor{gray}{\pm14.1}}$ \\
\midrule
\multicolumn{10}{c}{\textit{Open-source LLMs (Small)}} \\
\midrule
Qwen3.5-9B
& 20\% & 64$_{\textcolor{gray}{\pm41}}$ & 13.3$_{\textcolor{gray}{\pm29.8}}$
& \cellcolor{green100}100\% & 60\% & 40\%
& \cellcolor{lightblue2}\underline{4.68} & \cellcolor{lightblue3}\underline{46.0} & 38.3$_{\textcolor{gray}{\pm56.6}}$ \\
Llama-3.1-8B-Instruct
& \cellcolor{verylightgreen}\underline{40\%} 
& \underline{83$_{\textcolor{gray}{\pm45}}$} 
& \underline{30.6$_{\textcolor{gray}{\pm42.1}}$}
& \cellcolor{green100}100\% 
& \cellcolor{green80}80\% 
& 40\%
& \cellcolor{lightblue2}2.86 
& \cellcolor{lightblue3}35.6 
& \underline{45.9$_{\textcolor{gray}{\pm38.1}}$} \\
Llama-3-8B-Instruct
& 20\% & 68$_{\textcolor{gray}{\pm37}}$ & 16.1$_{\textcolor{gray}{\pm36.0}}$
& \cellcolor{green100}100\% & \cellcolor{green80}80\% & 20\%
& \cellcolor{lightblue1}1.38 & \cellcolor{lightblue3}26.2 & 26.7$_{\textcolor{gray}{\pm30.8}}$ \\
Gemma-4-E4B-it
& 0\% & 32$_{\textcolor{gray}{\pm1}}$ & 0.0$_{\textcolor{gray}{\pm0.0}}$
& \cellcolor{green100}100\% & 0\% & 0\%
& \cellcolor{verylightblue}0.19 & \cellcolor{verylightblue}0.6 & 0.0$_{\textcolor{gray}{\pm0.0}}$ \\
NVIDIA-Nemotron-Nano-9
& 0\% & 38$_{\textcolor{gray}{\pm12}}$ & 0.0$_{\textcolor{gray}{\pm0.0}}$
& \cellcolor{green100}100\% & 20\% & 0\%
& \cellcolor{lightblue1}1.91 & \cellcolor{lightblue3}24.8 & 9.6$_{\textcolor{gray}{\pm7.0}}$ \\
\midrule
Overall (ReAct)
& 13.0\% & 57$_{\textcolor{gray}{\pm33}}$ & 9.8$_{\textcolor{gray}{\pm26.8}}$
& \cellcolor{green100}100\% & \cellcolor{lightgreen}50.4\% & 17.4\%
& \cellcolor{lightblue2}3.59 & \cellcolor{lightblue2}15.7 & 27.3$_{\textcolor{gray}{\pm35.3}}$ \\
\midrule
\multicolumn{10}{c}{\textit{Other State-of-the-art Agent Frameworks}} \\
\midrule
Claude Code + Opus 4.7
& 20\% & 67.4$_{\textcolor{gray}{\pm37}}$ & 31.7$_{\textcolor{gray}{\pm30.9}}$
& \cellcolor{green100}100\% & \cellcolor{green80}80\% & 20\%
& 0.25 & \cellcolor{lightblue3}67.4 & 39.1$_{\textcolor{gray}{\pm43.2}}$ \\
Codex CLI + GPT-5.5
& \cellcolor{lightgreen}\underline{\textbf{60\%}} 
& \underline{\textbf{115$_{\textcolor{gray}{\pm31}}$}} 
& \underline{34.7$_{\textcolor{gray}{\pm32.9}}$}
& \cellcolor{green100}100\% 
& \cellcolor{green80}80\% 
& \cellcolor{lightgreen}60\%
& \cellcolor{verylightblue}\underline{0.88} 
& \cellcolor{lightblue3}\underline{\textbf{114.6}} 
& \underline{60.0$_{\textcolor{gray}{\pm13.5}}$} \\
OpenClaw + DeepSeek-V4
& 20\% & 55$_{\textcolor{gray}{\pm48}}$ & 14.8$_{\textcolor{gray}{\pm32.2}}$
& \cellcolor{green100}100\% & 40\% & 20\%
& \cellcolor{verylightblue}0.27 & \cellcolor{lightblue3}55.6 & 24.7$_{\textcolor{gray}{\pm29.9}}$ \\
\midrule
Overall (Other Agents)
& \cellcolor{verylightgreen}{33.3\%} 
& {79$_{\textcolor{gray}{\pm39}}$} 
& {27.1$_{\textcolor{gray}{\pm32.0}}$}
& \cellcolor{green100}{100\%} 
& \cellcolor{lightgreen}{66.7\%} 
& {33.3\%}
& \cellcolor{verylightblue}{0.47} 
& \cellcolor{lightblue3}{79.2} 
& {41.3$_{\textcolor{gray}{\pm28.9}}$} \\
\midrule
Overall
& \cellcolor{verylightgreen}{15.4\%} 
& {60$_{\textcolor{gray}{\pm35}}$} 
& {11.8$_{\textcolor{gray}{\pm27.8}}$}
& \cellcolor{green100}{100\%} 
& \cellcolor{lightgreen}{52.3\%} 
& {19.2\%}
& \cellcolor{verylightblue}{3.23} 
& \cellcolor{lightblue3}{23.0} 
& {28.9$_{\textcolor{gray}{\pm35.2}}$} \\
\bottomrule
\end{tabular}
}
\caption{Performance comparison between ReAct Agent (115 runs across all backbone models) and other agent frameworks. Full Survival \% shows the fraction of runs that complete the entire 132-month simulation horizon; Avg. Mon. shows mean ($  \pm  $ SD) duration survived across 5 runs per model; Score is the valuation score defined in Section~\ref{sec:eval}; 1st, 2nd, and 3rd Crisis Survival \% reports the proportion of runs surviving through each respective shock event; Tools/Mo records average tool usage per month per run; Actions show average number of effective (non-pass) actions per run; Raised -- cumulative capital raised across runs.
\underline{Underlining} denotes best result within each sub-category; \textbf{boldface} highlights best overall performance across categories.}
\label{tab:enterprise_results}
\end{table*}

\noindent{\textbf{Human baseline.}}
To contextualize agent performance against experienced human judgment, we included five finance experts with an average of over ten years of experience in finance as human baselines. Three are affiliated with current or former CFO organizations and have direct enterprise finance experience, and the remaining two specialize in banking and credit risk (details in Appendix~\ref{appendix_human_experts}). Their performance serves as a reference for interpreting agent results and validating that the environment admits effective strategies when guided by domain expertise.


\subsection{Results and Analysis}
\label{sec:results}



\noindent{\textbf{RQ1: Can LLM agents perform long-horizon resource allocation under uncertainty?}}
Table~\ref{tab:enterprise_results} reports results across 23 backbone models under ReAct and three SOTA agent frameworks. Under ReAct, only 13\% of trials survive the full 132-month horizon, and 12 out of 23 models never survive a single run. Failures cascade across economic crises: all models pass the first downturn, but only 50\% survive the second, and fewer than 18\% reach the third.
Only two configurations match the human survival rate of 60\%: DeepSeek-v4-pro under ReAct and Codex CLI with GPT-5.5. Yet even the best agent score of \$34.7M from Codex reaches only 7\% of the human baseline of \$476.7M, despite comparable tool usage, indicating that the bottleneck is not information access but the ability to convert information into well-timed commitments.
Model scale does not predict performance. Llama-3.1-8B-Instruct at 8B parameters achieves 40\% survival and \$30.6M, outperforming Llama-3.3-70B-Instruct that never survives. Grok-4.3 at 40\% outperforms Grok-4.20 at 0\%. DeepSeek-V4 at 60\% far exceeds DeepSeek-V3.1 at 0\%.

We find that framework choice also matters, but in unexpected ways. GPT-5.5 fails completely under ReAct at 0\% survival yet achieves 60\% with Codex, where it takes 114.6 actions versus 17.2 under ReAct and uses tools at just 0.88 per month. Codex appears to unlock a sustained engagement that the ReAct loop does not. DeepSeek-v4-pro tells the opposite story: it thrives under ReAct at 60\% survival with heavy tool use of 16.2 per month, but drops to 20\% under OpenClaw, where tool usage falls to 0.27 per month, as if the framework's harness cuts off the very observation strategy that makes the model effective. 
\begin{figure*}[htbp]
  \centering
  \includegraphics[width=\linewidth]{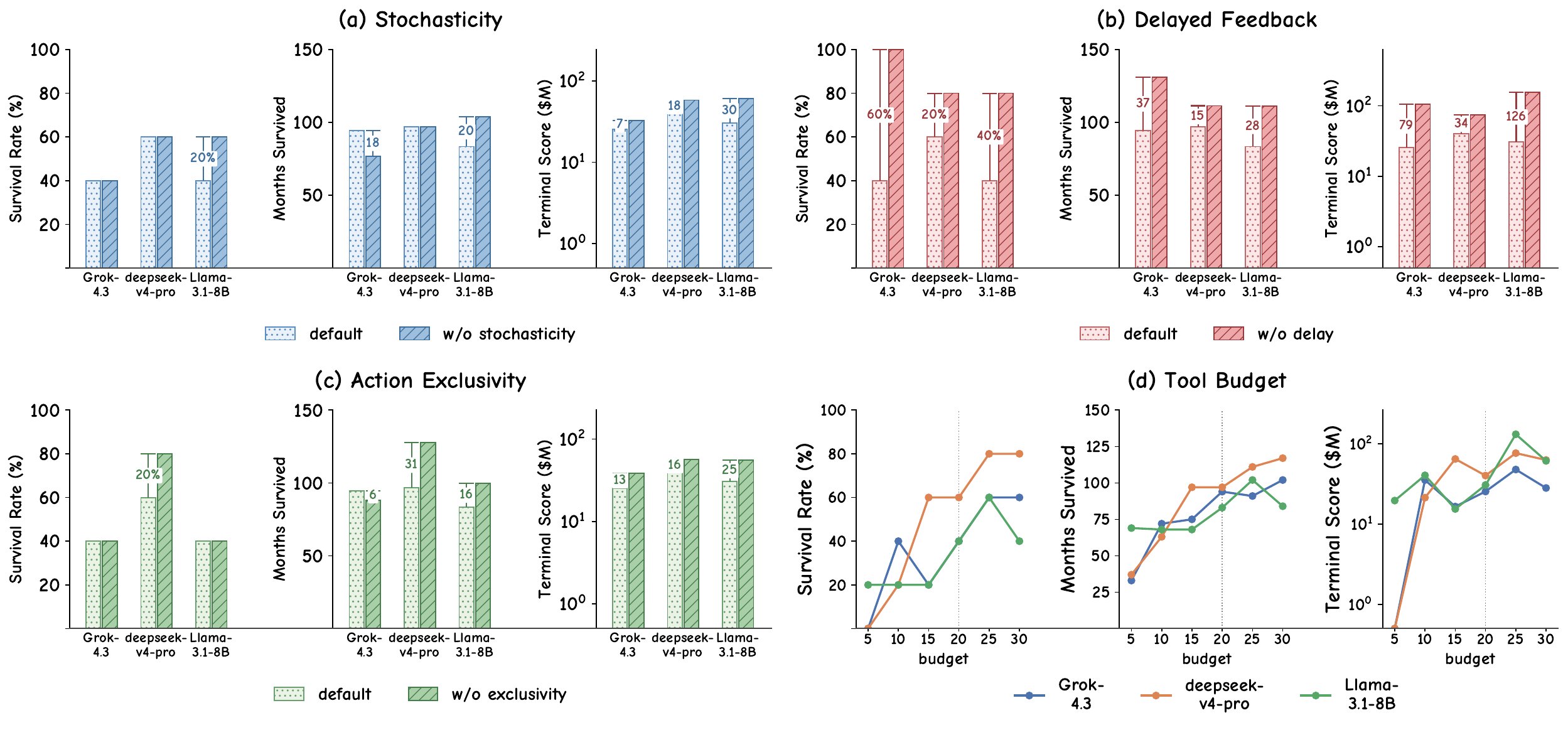}
  \caption{Ablation study across three backbone LLMs ($n{=}5$ runs per cell). Panels~(a), (b) and (c) each remove one design knob from the default arena: (a)~ environment stochasticity, (b)~delayed feedback, and (c)~action exclusivity. And panel~(d) sweeps the tool budget $B \in \{5,10,15,20\ (default),25,30\}$.}
  \label{fig:ablation_panels}
\end{figure*}

\noindent{\textbf{RQ2: Which environmental conditions contribute most to the difficulty?}}
To diagnose where agent performance breaks down, we systematically relax each environmental constraint on three representative models (Grok-4.3, DeepSeek-V4, Llama-3.1-8B-Instruct): including fundraising settlement delay, environmental stochasticity, per-step tool budget, and action exclusivity.

The results are shown in Figure~\ref{fig:ablation_panels}. We can see that \textbf{Delayed feedback} is the dominant factor. Removing the settlement delay yields the largest improvement across all metrics: Grok-4.3 rises from 40\% to 100\% survival, and Llama-3.1-8B-Instruct's terminal score jumps from \$31M to \$157M. Without delay, resource allocation reduces to a reactive task, the agent observes each outcome before committing again. With delay, effective allocation requires foresight: the agent must anticipate future risks and commit resources while prior outcomes remain unresolved. The magnitude of this improvement suggests that current models cannot allocate resources proactively in anticipation of future needs; instead, they defer action until a crisis has already materialized.
\textbf{Action exclusivity} has a moderate effect. Allowing agents to reconcile and fundraise in the same month improves terminal scores consistently (+40--82\%), but survival rates remain unchanged for two of three models. This dissociation is informative: agents know how to reconcile and how to fundraise, but when forced to choose one per step, they cannot reliably prioritize between them. The unchanged survival rates further confirm that prioritization alone does not explain the full difficulty, even without exclusivity, agents still fail to time their commitments correctly.
\textbf{Tool budget} also matters. All models collapse at budget 5, and budget 25 consistently outperforms the default of 20. Yet human experts achieve the highest performance with only 0.58 tools per month, far below any budget tested. This gap indicates that the bottleneck is not the amount of information available, but the ability to identify what matters and allocate observation resources accordingly. 
\textbf{Stochasticity} has the smallest effect: removing operational noise barely changes survival rates, suggesting that uncertainty amplifies failure but is not its root cause.
Together, these results show that current agents remain reactive rather than proactive. They respond to observed outcomes but fail to anticipate future needs, identify key information, and allocate limited resources effectively.



\begin{figure*}[t]
\centering
\includegraphics[width=\linewidth]{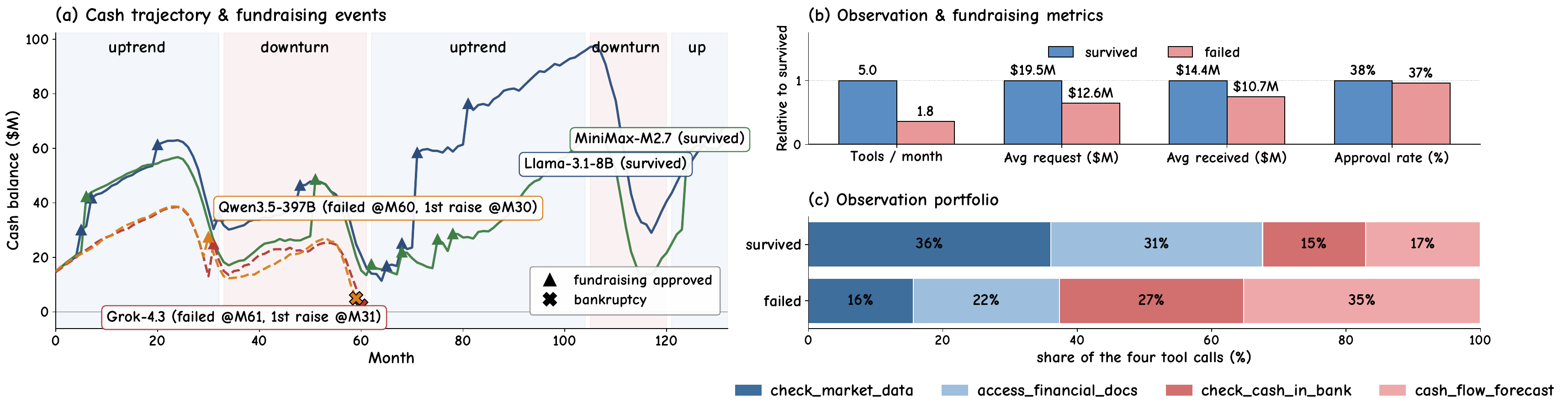}
\caption{{(a)} Cash trajectories of four representative ReAct trials. Shaded bands mark the macro uptrend and downturn periods. Markers show approved fundraising requests ($\blacktriangle$) and bankruptcy ($\times$). {(b)} Observation and fundraising metrics aggregated over 10 random survived and failed full-horizon ReAct trials, normalised so the survived value is 1.0. Bars from left to right: tool calls per month, average request size, average cash received per approval, fundraising approval rate. Absolute values are annotated above each bar. {(c)} Share of tool calls each group spends on four most used tools.}
\label{fig:failure_combined}
\end{figure*}

\noindent{\textbf{RQ3: How do agents fail at resource allocation?}}
We randomly sample 10 survived and 10 failed ReAct trials to identify where allocation breaks down. As shown in Figure~\ref{fig:failure_combined}, we identify three allocation failure modes that correspond to different layers of the resource allocation process.
First, failed runs act too late. As shown in Figure~\ref{fig:failure_combined}a, they allocate their first action slot to fundraising only after cash has peaked and begun to decline. By this point, the 1--6 month settlement delay makes it impossible to receive capital before insolvency. Survivors, by contrast, begin fundraising during the uptrend, responding to signals early and securing capital while conditions still appear favorable and building a buffer that helps absorb the subsequent downturn.
Second, failed agents do not achieve the optimal level of resource allocation needed to support efficient action -- failed agents use tools 2.8$\times$ less frequently and request smaller amounts when they do fundraise. As shown in Figure~\ref{fig:failure_combined}b, survivors average 5.0 tool calls per month versus 1.8 for failed runs, and request \$19.5M versus \$12.6M per attempt, receiving \$14.4M versus \$10.7M per approval. Approval rates are identical at 37--38\% across both survival and bankrupt trajectories, confirming that the gap is an agent-side allocation choice, not environmental bias.
Third, failed agents allocate their observation budget toward the wrong signals. As shown in Figure~\ref{fig:failure_combined}c, survivors spend 67\% of tool calls on external signals such as market data and financial documents, while failed runs spend 62\% on internal state such as cash balance and forecasts. By directing scarce observation resources inward, failed agents track what has already happened rather than what is about to change.
These findings show that current LLM agents still have a substantial gap in complex, dynamic environments: they cannot identify the right information to attend to, recognize critical decision points, or allocate scarce resources at the right time and scale.

\section{Conclusion}

We presented \textsc{EnterpriseArena}, a 132-month CFO simulator that evaluates LLM agents on long-horizon resource allocation under uncertainty. Experiments across 23 backbone LLMs and four agent frameworks show that current agents still struggle in this setting: only 15.4\% of trials survive the full horizon, model scale does not reliably predict success, and even the strongest configuration reaches just 7\% of the human expert baseline. A hierarchical pattern of failures, mis-allocated observation, mis-timed fundraising, and under-sized capital requests, suggests that current agents cannot sustain coherent strategies across sequences of binding decisions. Our findings establish long-horizon resource allocation as a distinct and challenging capability for LLM agents.

\section*{Limitations and Ethical Concerns}
Several limitations should be noted. Our environment remains a simulation and cannot capture extreme events like prolonged funding droughts or market freezes. We model AI agents rather than the multi-stakeholder hierarchy found in real organizations. See Appendix~\ref{limitation} for details.

\section*{Potential risks}

\paragraph{Potential Positive Impacts.}
By introducing a controlled environment for long-horizon enterprise decision-making, this work provides a systematic way to evaluate how LLM agents allocate scarce resources under uncertainty in an enterprise setting. It may support the development of more reliable planning strategies, improve benchmarking for real-world enterprise use cases, and encourage research into decision-making robustness in the long-term rather than short-term task performance.

\paragraph{Potential Negative Impacts.}
However, several risks remain. First, strong performance on \textsc{EnterpriseArena} could be misinterpreted as evidence that LLM agents are ready for real-world financial or operational deployment, despite the persistent gap between simulated and real-world complexity. Second, the benchmark may incentivize overfitting to specific simulator dynamics or evaluation metrics, leading to systems that perform well in controlled settings but fail under distributional shifts. Third, insights into agent weaknesses in resource allocation could be exploited to design adversarial scenarios that deliberately induce poor long-term decisions or resource exhaustion.

These findings should therefore be interpreted as diagnostic signals of current limitations, rather than indicators of deployment readiness.

\section*{Ethical considerations}
All experiments use publicly available models and datasets, and do not involve any personal, sensitive, or confidential company information.

\begin{ack}
Use unnumbered first level headings for the acknowledgments. All acknowledgments
go at the end of the paper before the list of references. Moreover, you are required to declare
funding (financial activities supporting the submitted work) and competing interests (related financial activities outside the submitted work).
More information about this disclosure can be found at: \url{https://neurips.cc/Conferences/2026/PaperInformation/FundingDisclosure}.

Do {\bf not} include this section in the anonymized submission, only in the final paper. You can use the \texttt{ack} environment provided in the style file to automatically hide this section in the anonymized submission.
\end{ack}

\bibliographystyle{plainnat}   
\bibliography{custom}          


\appendix

\section{Comparison of prior benchmarks}
\label{compare}

Table~\ref{tab:benchmark_comparison} compares \textsc{EnterpriseArena} with representative general-purpose and financial agent benchmarks along the four structural properties required to evaluate long-horizon resource allocation under uncertainty. These properties are not intended to measure whether a benchmark is broadly useful or difficult. Instead, they identify whether the benchmark places an agent in a setting where decisions consume scarce capacity, consequences unfold over delayed and partially observed trajectories, and future states are shaped by both exogenous changes and the agent's own prior commitments.

For general-purpose agent benchmarks, existing environments cover important aspects of agentic behavior, including web navigation, software engineering, tool use, workplace simulation, and long-horizon memory. However, most of these tasks evaluate whether an agent can complete externally specified goals in environments where feedback is relatively explicit and errors remain local or recoverable. Even when the task is long-horizon, such as software engineering or workplace simulation, the agent is usually not forced to allocate a shared scarce resource across competing operational needs. Similarly, trajectory-based memory benchmarks may require reasoning over extended histories, but the agent does not act in a closed loop where its decisions alter the future environment.

Financial agent benchmarks are closer to our target setting, but they also differ in important ways. Trading and investment benchmarks often involve long sequences of market-facing decisions, yet the core action is typically a reversible buy, sell, or hold decision, and the agent's choices do not consume a limited enterprise operating capacity that forecloses other actions. Judgment-oriented benchmarks evaluate financial analysis or recommendation quality, but their outputs do not become binding actions inside an evolving environment. Financial reasoning benchmarks test domain knowledge and tool use, but generally do not model an enterprise whose internal state, liquidity, and future constraints evolve as a consequence of the agent's prior commitments.

\textsc{EnterpriseArena} is designed to instantiate all four properties simultaneously. The agent operates under a hard action budget, where both operational actions and information-gathering tools consume scarce capacity. The horizon spans 132 monthly decision steps, requiring strategies that remain coherent across multiple macroeconomic regimes. Consequences are latent because the true financial state is only partially observable and the value of actions, such as fundraising, book closing, or holding capacity, may become clear only after substantial delay. Finally, the environment is non-stationary because macroeconomic conditions shift over time and internal enterprise dynamics depend on both exogenous signals and the agent's accumulated decisions. This combination distinguishes \textsc{EnterpriseArena} from prior benchmarks that test individual components of agentic behavior but do not jointly stress-test long-horizon resource allocation under uncertainty.

\begin{table*}[t]
\centering
\small
\setlength{\tabcolsep}{6pt}
\renewcommand{\arraystretch}{1.15}
\adjustbox{max width=\textwidth}{
\begin{tabular}{l l c c c c}
\toprule
\textbf{Benchmark} & \textbf{Domain} 
& \textbf{Resource Allocation} 
& \textbf{Long Horizon}
& \textbf{Latent Consequences}
& \textbf{Non-stationary} \\
\midrule
\rowcolor{groupgray}
\multicolumn{6}{l}{\textit{General-purpose agent benchmarks}} \\
\midrule
WebArena~\cite{zhou2023webarena} 
& Web Navigation 
& \xmarkc & \xmarkc & \xmarkc & \xmarkc \\
SWE-Bench-pro~\cite{deng2025swebenchproaiagents} 
& Software Engineering 
& \xmarkc & \cmarkc & \xmarkc & \xmarkc \\
AgentBench~\cite{liu2025agentbenchevaluatingllmsagents} 
& Multi-domain Tasks 
& \xmarkc & \xmarkc & \xmarkc & \xmarkc \\
TheAgentCompany~\cite{xu2025theagentcompanybenchmarkingllmagents} 
& Enterprise Workplace 
& \xmarkc & \cmarkc & \xmarkc & \xmarkc \\
AMA-Bench~\cite{zhao2026amabenchevaluatinglonghorizonmemory} 
& Trajectory Memory QA 
& \xmarkc & \cmarkc & \cmarkc & \xmarkc \\
\midrule
\rowcolor{groupgray}
\multicolumn{6}{l}{\textit{Financial agent benchmarks}} \\
\midrule
InvestorBench~\cite{li2025investorbench} 
& Investment Decision 
& \xmarkc & \cmarkc & \xmarkc & \xmarkc \\
AMA~\cite{qian2026agents} 
& Market Trading 
& \xmarkc & \cmarkc & \xmarkc & \xmarkc \\
AI-Trader~\cite{fan2025aitraderbenchmarkingautonomousagents} 
& Market Trading 
& \xmarkc & \cmarkc & \xmarkc & \xmarkc \\
STOCKBENCH~\cite{chen2025stockbench} 
& Stock Trading 
& \xmarkc & \cmarkc & \xmarkc & \xmarkc \\
Finance Agent Benchmark~\cite{bigeard2025financeagentbenchmarkbenchmarking} 
& Financial Research 
& \xmarkc & \xmarkc & \xmarkc & \xmarkc \\
FinGAIA~\cite{zeng2025fingaiachinesebenchmarkai} 
& Financial Reasoning 
& \xmarkc & \xmarkc & \xmarkc & \xmarkc \\
\midrule
\rowcolor{ourshighlight}
\textbf{EnterpriseArena (Ours)} 
& \textbf{Enterprise Financial Management} 
& \cmarkc & \cmarkc & \cmarkc & \cmarkc \\
\bottomrule
\end{tabular}
}
\caption{
Comparison with representative agent benchmarks along the four structural properties \textbf{whose simultaneous presence is required to stress-test long-horizon resource allocation under uncertainty}. \cmarkc~denotes that the property is core to the benchmark setting; \xmarkc~denotes that it is not present. Each property appears individually in prior work, but their co-occurrence is what existing benchmarks do not provide.
\textbf{Resource Allocation}: agents operate under hard budgets on both action capacity and observation capacity.
\textbf{Long Horizon}: the task involves sequential decision-making or reasoning over extended trajectories spanning many steps.
\textbf{Latent Consequences}: ground-truth state is hidden unless the agent spends scarce action capacity to reveal it, and action outcomes unfold over delays during which only partial signals are available.
\textbf{Non-stationary}: environment dynamics are stochastic and state-dependent, with transitions that shift across regimes and depend on the agent's prior actions.
}
\label{tab:benchmark_comparison}
\end{table*}

\section{Environment and Dataset}
\label{sec:appendix}

\subsection{Details of Environment Settings}
 
\subsubsection{Environment components}
\label{app_obs_section}
The \textsc{EnterpriseArena} environment consists of three layers that jointly shape the enterprise's state and the agent's decision context. The economic environment captures macroeconomic conditions and capital market signals. The industry environment provides sector-level benchmarks. The enterprise environment contains the firm's internal financial and operational records. Table~\ref{tab:observation_space} lists the specific indicators available in each layer.

\begin{table}[h]
\centering
\footnotesize
\setlength{\tabcolsep}{6pt}

\resizebox{0.7\linewidth}{!}{%
\begin{tabular}{lll}
\toprule
\textbf{Environment} & Indicator& \textbf{Observable Signals} \\
\midrule

\textit{Economic Environment} & Economic & GDP, CPI, unemployment \\
 & Debt Fundraising& Interbank rates, Bond yields \\
 & Equity Fundraising& VIX index, P/E ratios, revenues \\

\midrule
\textit{Industry Environment} & Growth &growth rate \\
 & Profitability& EBITDA margin \\

\midrule
\textit{Enterprise Environment} & Board Materials & Cap table, business overview \\
 & Financial Performance & Financial statements \\
 & Financial Operation& General ledger, vendor contracts \\
 & Liquidity & Cash balance \\

\bottomrule
\end{tabular}%
}

\caption{Observable signals in the EnterpriseArena environment.}
\label{tab:observation_space}
\end{table}

\subsubsection{Environment dynamic details}
\label{dynamic}
 
The environment evolves autonomously at each timestep to simulate the natural fluctuations of both internal enterprise operations and external economic conditions over time. Table~\ref{tab:env-dynamics} details the stochastic processes governing these dynamics. Internal enterprise indicators (e.g., margins, user growth) receive additive Gaussian perturbations monthly, reflecting the inherent unpredictability of business operations. External economic and industry indicators evolve through regime-dependent transition models calibrated on real-world historical data, capturing the shifting conditions across expansion, neutral, and recession phases. These dynamics are exogenous, they proceed regardless of the agent's actions, and the agent must adapt its strategy accordingly.

\begin{table*}[t]
\centering
\small
\resizebox{\linewidth}{!}{
\begin{tabular}{llll}
\toprule
\textbf{Indicator} & \textbf{What it controls} & \textbf{Transition Model} & \textbf{Parameters} \\
\midrule
\multicolumn{4}{l}{\textit{Enterprise Operational Dynamics (additive Gaussian, applied monthly)}} \\
\addlinespace[2pt]
Gross margin        & Revenue $\to$ gross profit conversion  & $x'_t = x_t + \epsilon,\ \epsilon \sim \mathcal{N}(0, \sigma^2)$, clipped $[10, 80]$  & $\sigma = 2.0$ \\
EBITDA margin       & Post-operating-cost retention           & $x'_t = x_t + \epsilon,\ \epsilon \sim \mathcal{N}(0, \sigma^2)$, clipped $[0, 60]$   & $\sigma = 1.5$ \\
User growth rate    & Customer base expansion rate            & $x'_t = x_t + \epsilon,\ \epsilon \sim \mathcal{N}(0, \sigma^2)$                       & $\sigma = 0.5$ \\
Collection rate     & Receivable-to-cash conversion           & $x'_t = x_0 + \epsilon,\ \epsilon \sim \mathcal{N}(0, \sigma^2)$, clipped $[0.85, 1.0]$ & $\sigma = 0.04$;\; $x_0 = 0.97$ (fixed) \\
\midrule
\multicolumn{4}{l}{\textit{External Economic \& Industry Dynamics (deterministic, exogenous)}} \\
\addlinespace[2pt]
Macroeconomic       & GDP, CPI, unemployment, interest rates & Fixed 132-month trajectory from historical data            & From Jan 2015 to Dec 2025, monthly frequency, anonymized \\
Equity market       & VIX, P/E ratio, P/S ratio              & Fixed 132-month trajectory from historical data            & From Jan 2015 to Dec 2025, monthly frequency, anonymized \\
\bottomrule
\end{tabular}
}
\caption{Stochastic and exogenous dynamics of the \textsc{EnterpriseArena} environment. \textit{Top:} enterprise operational indicators receive additive Gaussian noise each month (larger $\sigma$ means higher volatility; some are clipped to realistic ranges). 
\textit{Bottom:} external economic and industry indicators follow a fixed path derived from anonymized historical data and are exogenous to agent actions.}
\label{tab:env-dynamics}
\end{table*}

\begin{table*}[h]
\centering
\small
\resizebox{\textwidth}{!}{%
\begin{tabular}{llll}
\toprule
\textbf{Feedback} & \textbf{What it determines} & \textbf{Distribution} & \textbf{Key Driver} \\
\midrule
Funding outcome (equity) & Whether the equity raise succeeds & $\text{Bernoulli}(p_{\text{equity}}(t) \times 0.75^{n})$ & Market sentiment (VIX-driven); decays with prior rounds \\
Funding outcome (debt) & Whether the debt raise succeeds & $\text{Bernoulli}\big(p_{\text{debt}}(t) \times m_{\text{company}}(L)\big)$ & Credit conditions (SOFR-driven); penalized by leverage \\
\addlinespace[2pt]
Fill rate for capital raised & Raised capital from total requested amount & $\mathcal{U}(0.7,\, 1.0)$ & Randomly determined per attempt \\
Settlement delay & Months until funds are available & $\mathcal{U}_{\mathbb{Z}}(1,\, 6)$ & Randomly determined per attempt \\
\addlinespace[2pt]
Contract rate & Interest rate on the new debt & $r_t^{\text{debt}} = r_t + 500\,\text{bps} \times (L{-}0.5)^+$ & Macro yields $r_t$ + leverage spread; visible at request time \\
\bottomrule
\end{tabular}%
}
\caption{Fundraising feedback mechanics. $n$: number of prior successful equity rounds; $L$: debt-to-equity leverage ratio; $r_t = (\text{Tsy2Y}_t + \text{Baa}_t)/100$: base lending rate; $d$: settlement delay. Only the contract rate is visible to the agent at request time; other outcomes are revealed after $d$.}
\label{tab:fundraising}
\end{table*}

\subsubsection{Environment feedback for fundraising.}
\label{app:fundraising}
 
When the agent submits a fundraising request, the environment returns stochastic feedback that determines the outcome, see Table~\ref{tab:fundraising} for more details.
 
The central feedback is the funding outcome, a binary success or failure sampled as $\text{Bernoulli}(p_{\text{adj}})$, where $p_{\text{adj}} = p_{\text{macro}} \times m_{\text{company}}$. This factorization reflects the two forces that jointly determine whether a real-world fundraising attempt succeeds: external market conditions and the firm's own financial standing.
 
For equity and debt, these two components are driven by different factors. On the market side, equity uses $p_{\text{macro}}$ derived from equity market sentiment (VIX index), while debt uses $p_{\text{macro}}$ derived from credit market conditions (Federal Funds rate and Treasury yields). On the company side, equity applies a round-count decay $m_{\text{company}} = 0.75^{n}$ ($n$ = prior successful rounds), reflecting that investors become increasingly reluctant in later rounds due to dilution and valuation concerns. Debt applies a leverage penalty:
\begin{equation}
    m_{\text{company}} = \max\!\big(0,\; 1 - 1.5 \times \max(0,\; L - 0.5)\big),
\end{equation}
where $L$ is the debt-to-equity ratio. No penalty is applied when $L \leq 0.5$, and the probability decreases linearly until reaching zero at $L \approx 1.17$, reflecting that lenders are unwilling to extend credit to highly leveraged firms.
 
Conditional on success, the environment further determines how much capital is received, when it arrives, and at what cost. The fill rate $f \sim \mathcal{U}(0.7,\, 1.0)$ means only a portion of the requested amount may be funded, yielding $A_{\text{actual}} = f \cdot A_{\text{requested}}$. This reflects that real-world fundraising rarely fills the exact target, investors may partially commit due to risk appetite, portfolio constraints, or negotiation outcomes.

The raised capital is not immediately available. A settlement delay $d \sim \mathcal{U}_{\mathbb{Z}}(1,\, 6)$ determines how many months before the funds are deposited into the enterprise's cash balance. In practice, fundraising involves due diligence, legal review, and multi-party coordination that introduce unavoidable latency between commitment and receipt. This delay is a critical design choice: it prevents the agent from treating fundraising as an instant rescue mechanism and instead requires forward planning, the agent must anticipate liquidity needs months in advance and commit to fundraising before a cash crisis materializes.
 
For debt, the contract interest rate is set by market conditions at the time of settlement ($t{+}d$) rather than at the time of request. This means the settlement delay introduces a second layer of uncertainty: not only does the agent not know exactly when the funds will arrive, but it also cannot predict the exact borrowing cost it will face. A request initiated under favorable interest rate conditions may settle months later under less favorable ones, further complicating the agent's planning.
 

\section{Details of the curated dataset}
\label{app:dataset}
Table~\ref{tab:data_statistics} summarizes the 16 types of data and documents collected to construct the \textsc{EnterpriseArena} environment, spanning three layers. At the economic level, we collect macroeconomic indicators (GDP, CPI, unemployment), debt market signals (interbank rates, corporate and government bond yields), and equity market signals (VIX, P/E ratios, revenue multiples), all covering the full 132-month simulation period at monthly frequency from public sources including FRED, CBOE, and S\&P Global. At the industry level, we collect sector growth rates and profitability benchmarks (gross margin, EBITDA margin) derived from industry reports. At the company level, we use transformed and anonymized board materials, financial statements, and vendor contracts from public filings as the initialization of the simulated enterprise at $t=0$. Together, these three layers provide the data foundation for the environment dynamics described in Section~\ref{sec:env}.

\begin{table*}[h]
\centering
\footnotesize
\setlength{\tabcolsep}{4pt}
\resizebox{0.8\textwidth}{!}{
\begin{tabular}{lllccc}
\toprule
\textbf{Environment} & \textbf{Indicator Type} & \textbf{Data / Documents} & \textbf{Span} & \textbf{Freq.} & \textbf{Reference} \\
\midrule
Economic & Economic Indicator & GDP & 132 mo. & Monthly & FRED \\
 & Economic Indicator & CPI & 132 mo. & Monthly & FRED \\
 & Economic Indicator & Unemployment Rate & 132 mo. & Monthly & FRED \\
 & Debt Fundraising Availability & Interbank Rates (SOFR / Fed Funds) & 132 mo. & Monthly & FRED \\
 & Debt Fundraising Cost & Corporate Bond Yields (Baa OAS) & 132 mo. & Monthly & FRED \\
 & Debt Fundraising Cost & Gov. Bond Yields (2Y, 5Y, 10Y, 30Y) & 132 mo. & Monthly & FRED \\
 & Equity Fundraising Availability & VIX Index & 132 mo. & Monthly & CBOE \\
\hline
Industry & Growth Indicator & Industry User Growth & 132 mo. & Monthly & Public Industry reports\footnote{\label{fn:industry}\url{https://www.weforum.org/publications/the-future-of-global-fintech-2025/}, \url{https://us-go.experian.com/2025-state-of-fintech-report}} \\
\addtocounter{footnote}{-1}
& Valuation Indicator & Revenue Multiples & 132 mo. & Monthly & Public Industry reports\footnotemark \\
& Profitability Indicator & Industry Gross Margin & 132 mo. & Monthly & Public Industry reports\footnotemark \\
& Profitability Indicator & Industry EBITDA Margin & 132 mo. & Monthly & Public Industry reports\footnotemark \\
\hline
Company & Board Materials & Business Overview & t=0 & t=0 & PayPal 10-K \\
 & Board Materials & Capitalization Table & t=0 & t=0 & PayPal 10-K \\
 & Financial Performance & Financial Statements & t=0 & t=0 & PayPal 10-K \\
 & Financial Operations & Vendor Contracts & t=0 & t=0 & PayPal 10-K \\
 
\bottomrule
\end{tabular}
}
\caption{Foundation data collected for \textsc{EnterpriseArena}.}
\label{tab:data_statistics}
\end{table*}

\section{Evaluation}
\label{app:models}

We evaluate 23 LLMs instantiated as agents across four categories covering both proprietary and open-source ecosystems, with model scales ranging from small (8B) to very large MoE architectures. As summarized in Table~\ref{tab:models}, closed-source frontier models include GPT-5.5 and GPT-5.4~\cite{singh2025openaigpt5card,openai2026gpt54}, Gemini 3.1 Pro~\cite{google2026gemini31pro}, Claude Opus 4.7~\cite{anthropic2026claudeopus47systemcard} and Claude Haiku 4.5~\cite{anthropic2025haiku45}, and Grok 4.20~\cite{xai2026grok420reasoning} and Grok 4.3~\cite{xai2026grok43}. Large open-source models cover both dense and mixture-of-experts architectures: GLM-5 and GLM-5.1~\cite{glm5team2026glm5vibecodingagentic}, Qwen3.5-397B-A17B~\cite{qwen3.5}, DeepSeek-V3.1~\cite{liu2024deepseek} and DeepSeek-V4~\cite{deepseekai2026deepseekv4}, MiniMax-M2.7~\cite{minimax2026m27release}, and Llama-3.3-70B-Instruct~\cite{grattafiori2024llama}. Medium-scale models include Mistral-Small-24B-Instruct~\cite{mistral2025small31}, Mixtral-8x7B-Instruct~\cite{jiang2024mixtralexperts}, Gemma-4-31B~\cite{googledeepmind2026gemma4modelcard}, and Qwen3.5-35B-A3B~\cite{qwen3.5}. Small-scale models include Qwen3.5-9B~\cite{qwen3.5}, Llama-3.1-8B-Instruct and Llama-3-8B-Instruct~\cite{grattafiori2024llama}, Gemma-4-E4B-it~\cite{googledeepmind2026gemma4modelcard}, NVIDIA-Nemotron-Nano-9B~\cite{nvidia2025nvidianemotronnano2}. 

\begin{table*}[t]
\centering
\scriptsize
\resizebox{\textwidth}{!}{
\begin{tabular}{l l l l l}
\toprule
Category & Model & Developer & Parameters & Repository \\
\midrule

\multirow{7}{*}{Closed-source frontier}
& GPT-5.5~\cite{singh2025openaigpt5card} 
& OpenAI 
& Undisclosed 
& gpt-5.5-2026-04-23 \\

& GPT-5.4~\cite{openai2026gpt54} 
& OpenAI 
& Undisclosed 
& gpt-5.4-2026-03-05 \\

& Gemini 3.1 Pro~\cite{google2026gemini31pro} 
& Google DeepMind 
& Undisclosed 
& gemini-3.1-pro-preview \\

& Claude Opus 4.7~\cite{anthropic2026claudeopus47systemcard} 
& Anthropic 
& Undisclosed 
& claude-opus-4-7 \\

& Claude Haiku 4.5~\cite{anthropic2025haiku45} 
& Anthropic 
& Undisclosed 
& claude-haiku-4-5 \\

& Grok 4.20~\cite{xai2026grok420reasoning} 
& xAI 
& Undisclosed 
& grok-4.20-0309-reasoning \\

& Grok 4.3~\cite{xai2026grok43} 
& xAI 
& Undisclosed 
& grok-4.3 \\

\midrule

\multirow{7}{*}{Large-scale open-source}
& GLM-5~\cite{glm5team2026glm5vibecodingagentic} 
& Zhipu AI 
& 754B 
& zai-org/GLM-5 \\

& GLM-5.1~\cite{glm5team2026glm5vibecodingagentic} 
& Zhipu AI 
& 754B 
& zai-org/GLM-5.1 \\

& Qwen3.5-397B-A17B~\cite{qwen3.5} 
& Alibaba 
& 397B
& Qwen/Qwen3.5-397B-A17B \\

& DeepSeek-V3.1~\cite{liu2024deepseek} 
& DeepSeek AI 
& 671B 
& deepseek-ai/DeepSeek-V3.1 \\

& DeepSeek-V4~\cite{deepseekai2026deepseekv4} 
& DeepSeek AI 
& 862B 
& deepseek-ai/DeepSeek-V4-Pro \\

& MiniMax-M2.7~\cite{minimax2026m27release} 
& MiniMax 
& 229B 
& MiniMaxAI/MiniMax-M2.7 \\

& Llama-3.3-70B-Instruct~\cite{grattafiori2024llama} 
& Meta AI 
& 70B 
& meta-llama/Llama-3.3-70B-Instruct \\

\midrule

\multirow{4}{*}{Medium-scale open-source}
& Mistral-Small-24B~\cite{mistral2025small31} 
& Mistral AI 
& 24B 
& mistralai/Mistral-Small-24B-Instruct-2501 \\

& Mixtral-8x7B~\cite{jiang2024mixtralexperts} 
& Mistral AI 
& 46.7B 
& mistralai/Mixtral-8x7B-Instruct-v0.1 \\

& Gemma-4-31B~\cite{googledeepmind2026gemma4modelcard} 
& Google DeepMind 
& 31B 
& google/gemma-4-31B \\

& Qwen3.5-35B-A3B~\cite{qwen3.5} 
& Alibaba 
& 35B
& Qwen/Qwen3.5-35B-A3B \\

\midrule

\multirow{5}{*}{Small-scale open-source}
& Qwen3.5-9B~\cite{qwen3.5} 
& Alibaba 
& 9B 
& Qwen/Qwen3.5-9B \\

& Llama-3.1-8B-Instruct~\cite{grattafiori2024llama} 
& Meta AI 
& 8B 
& meta-llama/Llama-3.1-8B-Instruct \\

& Llama-3-8B-Instruct~\cite{grattafiori2024llama} 
& Meta AI 
& 8B 
& meta-llama/Meta-Llama-3-8B-Instruct \\

& Gemma-4-E4B-it~\cite{googledeepmind2026gemma4modelcard} 
& Google DeepMind 
& 4.5B effective 
& google/gemma-4-E4B-it \\

& NVIDIA-Nemotron-Nano-9B~\cite{nvidia2025nvidianemotronnano2} 
& NVIDIA 
& 9B 
& nvidia/NVIDIA-Nemotron-Nano-9B-v2 \\

\bottomrule
\end{tabular}
}
\caption{Large language models evaluated along the ReAct axis. The suite covers 23 models across closed-source frontier systems and open-weight models at large, medium, and small scales. Parameter counts are reported when publicly available; otherwise, they are marked as undisclosed.}
\label{tab:models}
\end{table*}










To provide broad coverage of the open-source model landscape, we further evaluate four large-scale open-source models: GLM-5~\cite{glm5team2026glm5vibecodingagentic}, Qwen3.5-397B~\cite{qwen3.5}, DeepSeek-V3.1~\cite{liu2024deepseek}, and Llama-3.3-70B-Instruct~\cite{grattafiori2024llama}. These models span dense and mixture-of-experts (MoE) architectures with parameter sizes ranging from tens to hundreds of billions. In addition, we include two medium-scale open-source models, Mistral-Small-24B~\cite{mistral2025small31} and Mixtral-8x7B~\cite{jiang2024mixtralexperts}, which provide a balance between capability and computational efficiency. Finally, we evaluate two small-scale open-source models, Qwen3.5-9B~\cite{qwen3.5} and Llama-3-8B-Instruct~\cite{grattafiori2024llama}, representing lightweight models commonly used in resource-constrained deployments.

This model selection allows us to systematically analyze agent performance across different model families, openness levels, and parameter scales.




\section{Failure cases}
\label{app:case_study}
 
We examine two representative failure trajectories to illustrate the behavioral patterns identified above. 
 
\begin{figure*}[t]
    \centering
    \includegraphics[width=\textwidth]{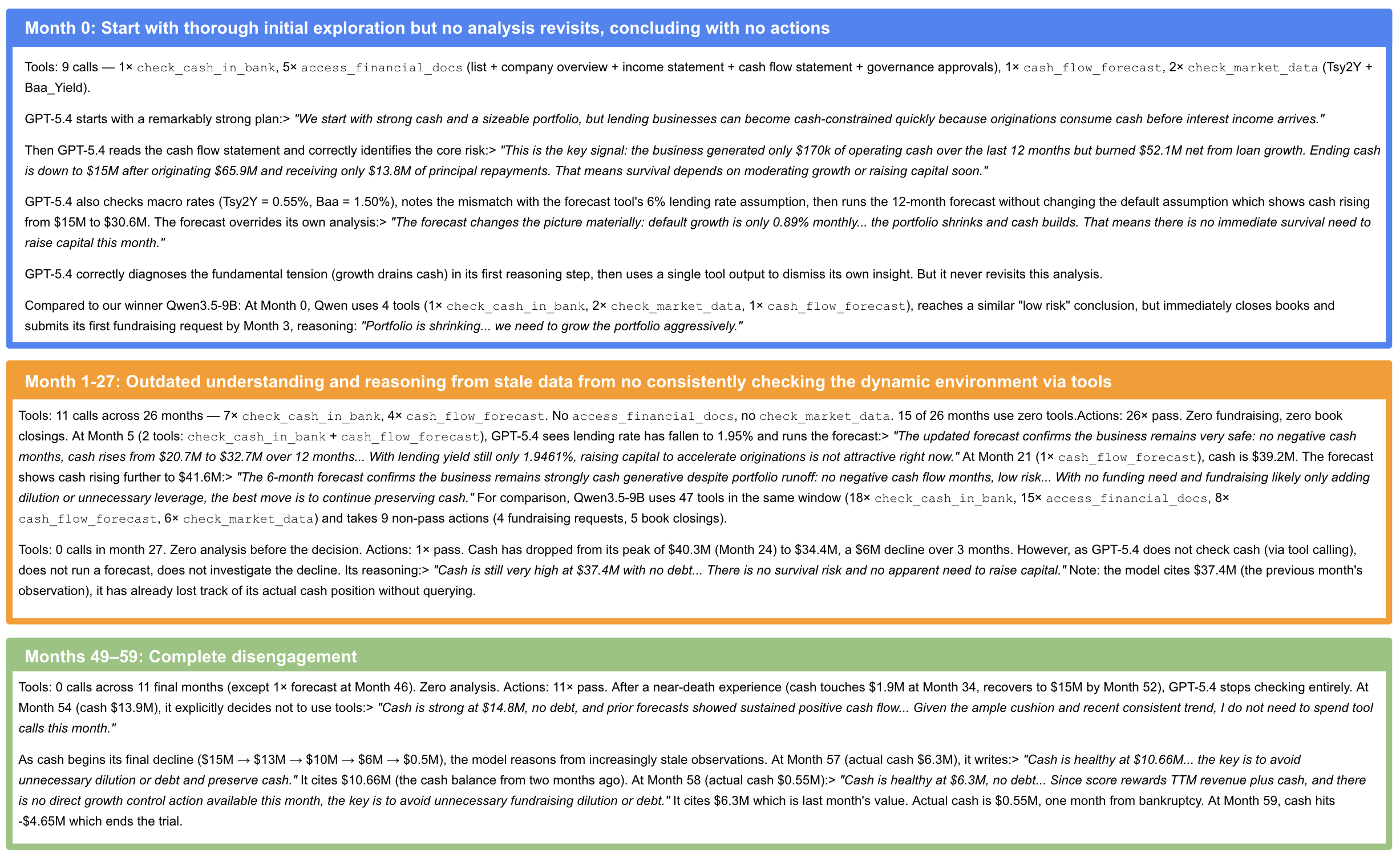}
    \caption{Failure Analysis 1: GPT-5.4}
    \label{fig:failure_analysis_gpt}
\end{figure*}

Figure~\ref{fig:failure_analysis_gpt} presents the trajectory of GPT-5.4. The agent performs an initial exploration at month 0 but never revisits its analysis or takes action. It reasons from increasingly stale data for months 1--27, then disengages entirely until cash runs out. Its 99.1\% pass rate reflects not poor judgment but a failure to maintain engagement with a dynamic environment.
 
\begin{figure*}[t]
    \centering
    \includegraphics[width=\textwidth]{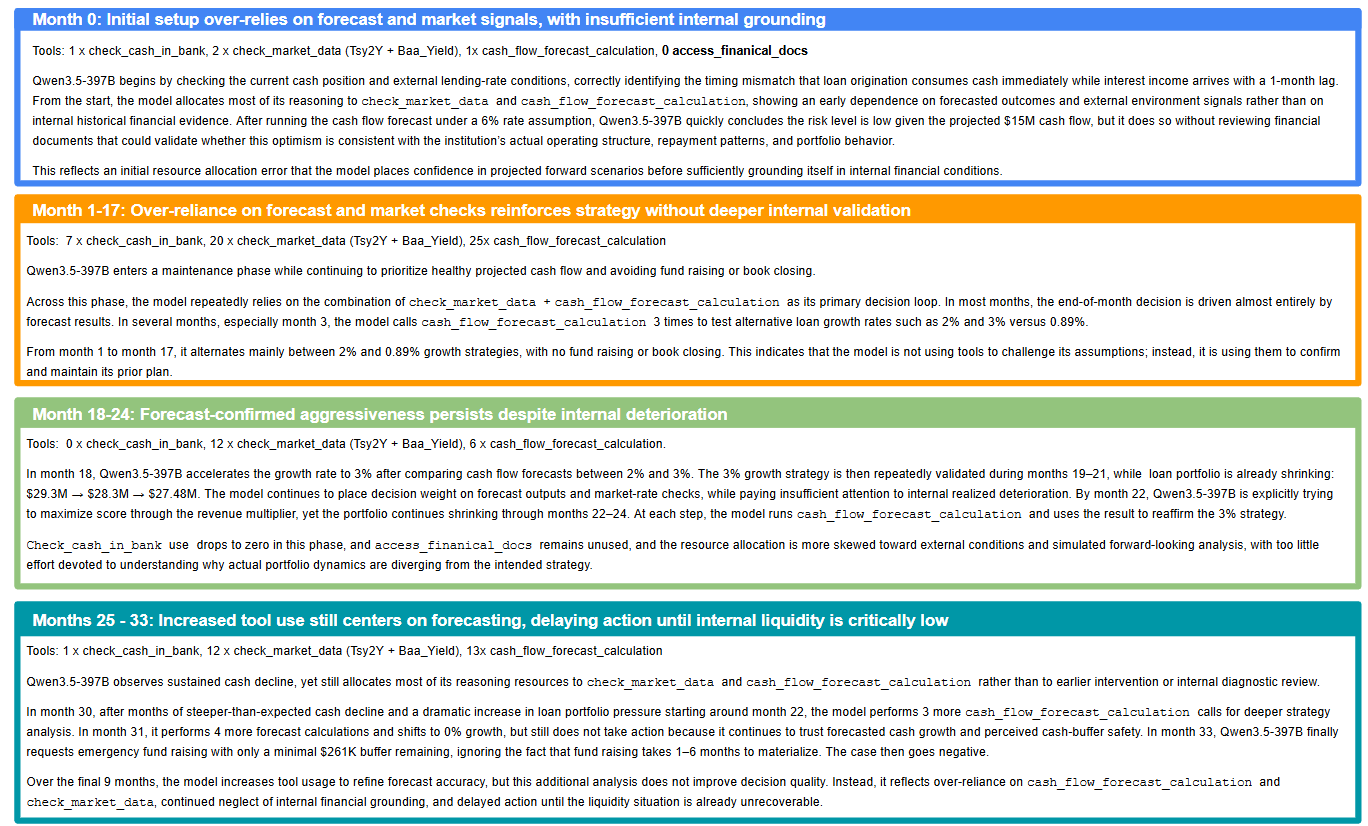}
    \caption{Failure Analysis 2: Qwen3.5-397B}
    \label{fig:failure_analysis_qwen}
\end{figure*}

Figure~\ref{fig:failure_analysis_qwen} presents the trajectory of Qwen3.5-397B, The model heavily uses forecasting and market tools throughout, yet almost never closes books (\textbf{book\_closing} 0.0\%) or acts on its analysis. It forms an optimistic plan at month 0 without checking internal records, reinforces it with repeated forecasts through months 1--24, and delays fundraising until cash is nearly exhausted. Extensive tool usage without timely action proves no better than complete inaction.

\section{Prompts}
\label{app:prompt}
Here is the prompt used for the agent.

\begin{tcolorbox}[promptbox,title={System Prompt for the CFO Agent}]
\ttfamily\small
You are an AI CFO (Chief Financial Officer) agent managing a company. Your primary objective is to ensure long-term solvency by maintaining a positive cash position at all times.

\medskip
\textbf{\#\# Available Tools}

\{tool\_descriptions\}

\medskip
\textbf{\#\# Memory Tools (persistent notepad)}

Important: Your conversation history resets every month. The only way to carry information across months is through your notepad. Your 5 most recent notes are automatically shown in each month's observation, but you can store many more and search them.

- save\_note(content, tags): Save a note to your notepad. Use tags to organize (e.g., ``cash'', ``fundraising'', ``macro'', ``strategy'').

- recall\_notes(query, tags, limit): Retrieve notes by keyword or tag. Without filters, returns the most recent notes.

\medskip
\textbf{\#\# Available Actions (state-changing)}

\{action\_descriptions\}

\medskip
\textbf{\#\# Rules}

- You have a tool budget of \{tool\_budget\} tool calls per month. You may use anywhere from 0 to \{tool\_budget\} tools in a single month.

- Memory operations (save\_note, recall\_notes) do NOT count toward the tool budget.

- Each month must end with exactly one action (fund\_raising\_request, book\_closing, or pass). You can take an action at any time, and you do not need to use any tools first.

- There is no prescribed order. You decide what information you need (if any) and when to act.

\medskip
\textbf{\#\# Response Format}

Think step by step, then output a JSON block:

For tools:

\{"type": "tool", "name": "tool\_name", "params": \{...\}\}

For actions:

\{"type": "action", "name": "action\_name", "params": \{...\}\}

For memory:

\{"type": "memory", "operation": "save\_note", "content": "...", "tags": ["tag1"]\}

\{"type": "memory", "operation": "recall\_notes", "query": "...", "tags": ["tag1"]\}

Always explain your reasoning before the JSON block.

\medskip
Current month: \{current\_month\}

Current date: \{current\_date\}
\end{tcolorbox}

\section{Human experts}
\label{appendix_human_experts}

\subsection{Human experts qualifications}
The first expert is a Chartered Financial Analyst (CFA) with over eight years of experience in enterprise finance. They have worked as a finance manager at leading technology and fintech companies as well as top-tier consulting firms. They hold a Master of Science in Finance from a prestigious U.S. university and a Bachelor of Economics with a major in Finance from a top university in China. This background gives them a strong dual perspective combining rigorous financial principles with extensive in-house and consulting expertise.

The second expert is a senior finance professional with over fourteen years of experience in corporate financial management and accounting operations. They have led financial planning and analysis, revenue accounting, and fiscal strategy at major global technology companies. They hold an MBA from a leading U.S. business school and a Bachelor’s degree from a renowned public research university. Their deep expertise in navigating the operational complexities of large-scale enterprises ensures a strong foundation in high-level corporate governance and disciplined resource allocation.

The third expert, currently working as a principal analyst at a major U.S. financial institution, holds a master’s degree in Business Analytics from a leading Ivy League university and a bachelor’s degree in Statistics and Economics. They have a solid educational background in accounting, finance, and strategic planning. Their experience includes research on large language models (LLMs), financial data analysis, and economics, enabling precise annotation of complex financial information.

The fourth expert is a finance and credit risk specialist with over eight years of experience at the intersection of commercial underwriting, regulatory risk management, and financial technology. They previously served as a credit analyst at a prominent U.S. commercial bank and advanced to Senior Consultant at a Big Four professional services firm’s Financial Services Risk Management practice. In this role, they have led major credit policy overhauls and conducted independent loan reviews on portfolios exceeding \$1 billion across commercial, real estate, REIT, securities-based lending, and structured finance.

The fifth expert is a finance data services leader with over ten years of experience building production-grade data systems and analytics solutions in the fintech and financial services sector. They spent more than four years at a fintech company focused on accessible credit and financial inclusion, where they led critical data initiatives supporting credit underwriting, risk modeling, data-driven lending decisions, and overall financial product strategy. Prior to that, they developed predictive analytics and modeling solutions with direct applications in financial risk assessment, customer finance, and credit operations.

\subsection{Expert Validation on Experiment Design}
\label{human-valid}
Two experts with extensive finance experience verified the financial consistency of both intermediate outcomes and the full trajectories generated by \textsc{EnterpriseArena} under standard accounting principles~\cite{securities2008topic,toerner2009guide}. Their review confirmed cross-statement consistency and realistic evolution of cash flow, equity, debt, and retained earnings over time. Specifically, the experts verified zero-error reconciliation across the balance sheet, income statement, and cash flow statement; and realistic volatility in gross and EBITDA margins under the calibrated stochastic processes. In quantitative spot-checks of 5 randomly sampled trajectories, all samples achieved 100\% reconciliation accuracy.

\subsection{Human baseline evaluation interface}
\begin{figure*}[t]
    \centering
    \includegraphics[width=\textwidth]{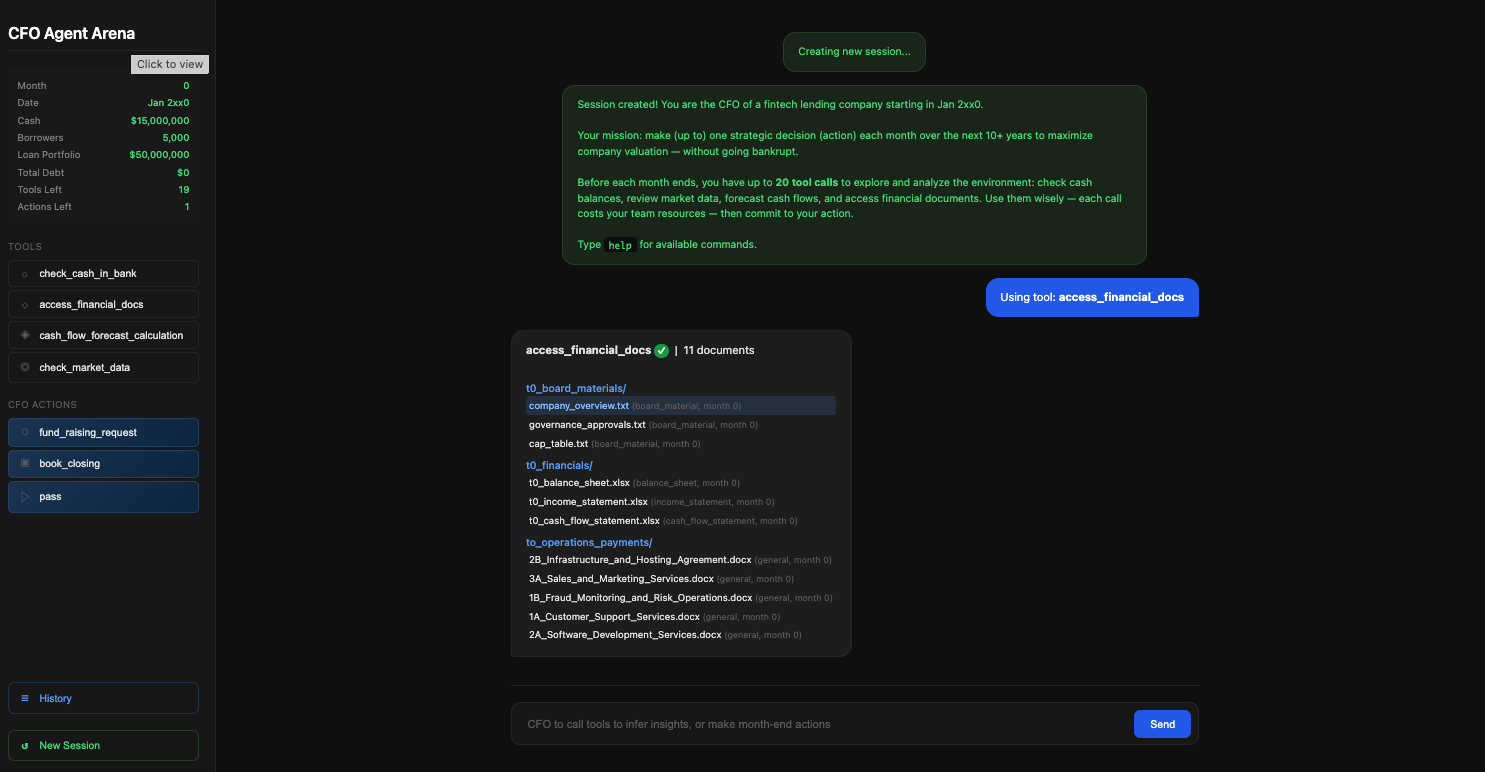}
    \caption{Human Baseline Evaluation Interface}
    \label{fig:human_ui}
\end{figure*}

Figure \ref{fig:human_ui} presents the interactive interface for experts to navigate the environment and complete the task.

\subsection{Instructions given to human experts}
Experts were given the interface shown in Figure~\ref{fig:human_ui} and received the full instructions as the prompt shown in Section~\ref{app:prompt} (the same system prompt given to the LLM agents). Experts were shown how to navigate the environment by performing analysis via tool calls and executing actions. Experts were informed that the simulation has no real financial risk. Furthermore, all participants were briefed that their performance results would serve as a human baseline for the study.

\subsection{Consent}
All experts provided explicit informed consent prior to participation. No personally identifiable information was collected or stored.

\subsection{Compensation}
The five experts participated on a voluntary basis with no monetary compensation provided.

\section{Experiments compute resources}
All experiments were conducted using a combination of local hardware (CPU/GPU) and API-based inference, with the latter being the primary compute source. Each run takes approximately 15–45 minutes per model, depending on task completion length, API latency, and local execution speed. Across all evaluated models (23 LLMs), agent frameworks, and both main experiments and ablations, the total runtime is approximately 70 hours. The overall API cost is estimated at around \$2000. No distributed systems or specialized compute infrastructure were used.

\section{Use of LLM}
We acknowledge that LLM-assisted tools were used during writing and editing (e.g., grammar checking, phrasing refinement, and formatting suggestions). These tools were not used to generate research ideas, experimental results, model outputs, or claims, and all technical content, analyses, and conclusions were designed, verified, and interpreted by the authors.

\section{Limitations}
\label{limitation}

First, while our environment uses expert-validated rules, anonymized public historical data and anonymized public financial materials, and stochastic dynamics calibrated to real-world volatility, it remains a simulation. Real-world enterprise finance environments feature "black swan" events and human irrationality that may not be fully captured by stochastic Gaussian noise or regime-dependent transitions. Similarly, the partial observability and delayed feedback mechanism (e.g. the 1--6 month stochastic settlement delay for fundraising) , while realistic, does not capture extreme corner cases like multi-year funding droughts or total market freezes.

Second, the current work utilizes a single-agent decision-making structure. In practice, resource-allocation decisions may occur within a multi-agent organizational hierarchy that includes the Board of Directors, multiple C-suites, department heads, external auditors, and other stakeholders whose objectives may conflict and require negotiation, coordination, and governance oversight.

Third, our empirical evaluation is initial, and remains limited in scope. Our current results are based primarily on the ReAct framework across 23 backbone LLMs, three additional agentic frameworks, and ablation studies for leading performers. We plan to expand the scope of our evaluation to include a broader range of state-of-the-art models and diverse agentic configurations in future work to ensure the generalizability and robustness of our findings.

Additionally, long-horizon evaluation over 132 monthly steps with repeated LLM reasoning and tool calls incurs computational cost. While feasible for offline strategic benchmarking, scaling to higher-frequency operations or very large organizations would require further optimizations such as model distillation, parallelization, or hierarchical planning.

\newpage

\end{document}